\documentclass{article} 
\usepackage{iclr2027_conference,times}

\usepackage{amsmath,amsfonts,bm}

\def\eqref#1{equation~\ref{#1}}

\def\1{\bm{1}}

\DeclareMathAlphabet{\mathsfit}{\encodingdefault}{\sfdefault}{m}{sl}
\SetMathAlphabet{\mathsfit}{bold}{\encodingdefault}{\sfdefault}{bx}{n}

\usepackage[utf8]{inputenc} 
\usepackage[T1]{fontenc}    
\usepackage{hyperref}       
\usepackage{url}            
\usepackage{booktabs}       
\usepackage{amsfonts}       
\usepackage{nicefrac}       
\usepackage{microtype}      
\usepackage{xcolor}         
\usepackage{bm,bbm,dsfont}
\usepackage{color}
\usepackage{colortbl}
\usepackage{soul}
\usepackage[most]{tcolorbox}
\tcbuselibrary{skins}
\tcbuselibrary{breakable}
\usepackage{CJK}
\usepackage{adjustbox}
\usepackage[fixed]{fontawesome5}

\usepackage{times}
\usepackage{latexsym}
\usepackage{multirow}
\usepackage{pifont}
\usepackage{amsmath, amsfonts}
\usepackage{algorithm}
\usepackage{algorithmicx}
\usepackage{algpseudocode}
\usepackage{tikz}

\usepackage{amsmath}
\usepackage{graphicx}
\usepackage{multirow}
\usepackage{multicol}
\usepackage{caption}
\usepackage{subcaption}
\usepackage{enumitem}
\usepackage{float}
\usepackage{makecell}
\usepackage{tabu}

\usepackage[T1]{fontenc}
\usepackage[utf8]{inputenc}
\usepackage{longtable}
\usepackage{color}
\usepackage{balance}
\usepackage{epigraph}
\definecolor{uclablue}{rgb}{0.15, 0.45, 0.68}
\hypersetup{
    breaklinks,
    citecolor=uclablue,
    colorlinks=true,
}

\definecolor{mygray}{gray}{0.6}

\usepackage{cleveref}
\usepackage{bbm}
\usepackage{wrapfig}

\newcommand{\bench}{$\Delta\tau$-Bench}
\newcommand{\ours}{\textsc{Steward}}

\definecolor{my_green}{RGB}{51,102,0}
\definecolor{my_red}{RGB}{204, 0, 0}
\definecolor{my_purple}{RGB}{160, 43, 147}
\definecolor{my_blue}{RGB}{15, 158, 213}
\definecolor{darkgrey}{rgb}{0.53,0.53,0.53}
\definecolor{mygrey}{rgb}{0.9,0.9,0.9}
\definecolor{circlered}{rgb}{252,100,84}

\definecolor{th_green}{rgb}{103,159,108}
\definecolor{code}{rgb}{59,94,176}
\definecolor{answer}{rgb}{182,27,25}
\definecolor{interp}{rgb}{0.53,0.53,0.53}
\newcommand{\coloredcircle}[2][red]

\definecolor{ResourcePink}{HTML}{EC008C}
\definecolor{GitHubNavy}{HTML}{101970}

\newcommand{\projectlinks}{%
  Code: \mbox{\href{https://github.com/zjunlp/delta-tau-bench}{%
    \textcolor{GitHubNavy}{\faGithub}\hspace{0.35em}%
    \textcolor{ResourcePink}{zjunlp/delta-tau-bench}}},%
  \hspace{1.2em}%
  Dataset: \mbox{\href{https://huggingface.co/datasets/zjunlp/delta-tau-bench}{%
    \raisebox{-0.18em}{%
      \includegraphics[height=1.2em]{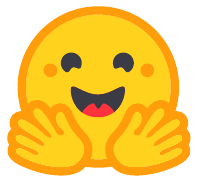}}%
    \hspace{0.35em}%
    \textcolor{ResourcePink}{zjunlp/delta-tau-bench}}}%
}

\title{\bench: Pushing Agents toward a Moving Capability Frontier}

\iclrfinalcopy

\author{
    Shuofei Qiao$^{\spadesuit\diamondsuit}$\footnotemark[1]~,
    Xuanguang Pan$^{\spadesuit}$\thanks{$\quad$ Equal Contribution.}~,
    Yijun Chen$^{\spadesuit}$,
    Wendi Feng$^{\spadesuit}$,
    Bin Wu$^\diamondsuit$, \\
    \textbf{ Runnan Fang$^{\spadesuit}$,
    Yuqi Zhu$^{\spadesuit}$,
    Shumin Deng$^{\spadesuit}$\footnotemark[2]~,
    Emine Yilmaz$^\diamondsuit$\thanks{$\quad$ Corresponding Author.}}\\
    $^\spadesuit$Zhejiang University ~
    $^\diamondsuit$University College London \\
    \fontsize{10.2pt}{0.1\baselineskip}\selectfont \texttt{\{shuofei,231sm\}@zju.edu.cn,} ~ \texttt{emine.yilmaz@ucl.ac.uk}
}

\begin{document}

\maketitle

\vspace{-.6cm}
\begin{abstract}
In this paper, we introduce {\bench}, an executable agent benchmark that models the structural evolution of large tool ecosystems. In {\bench}, independently evolving services continually reshape how tasks can be accomplished: familiar tools may disappear or be replaced, while more efficient alternatives emerge amid overlapping functionality and noisy documentation. Against this moving capability frontier, we evaluate whether agents can proactively discover efficient tool-use paths and promptly revise their accumulated procedural knowledge, bringing together demands that existing benchmarks fail to examine. {\bench} comprises 1,588 validated multi-turn tasks across 42 domains and six successive ecosystem states generated through capability-preserving transformations. Our evaluation shows that even the strongest evaluated model or harness achieves only 42.3\% task success, while initial gains from existing skill-learning methods can not consistently translate into better adaptation as the ecosystem evolves. These findings motivate {\ours}, an ecosystem manager that jointly manages tools and procedural knowledge through incremental skill maintenance and dynamic provisioning during task execution. Experiments show that {\ours} sustains higher task success across ecosystem changes with lower skill-maintenance overhead, while improving both recovery from disrupted workflows and adoption of more efficient alternatives\footnote{\projectlinks}.
\end{abstract}

\section{Introduction}
Large Language Model (LLM) agents are reshaping how users interact with software, enabling them to delegate goals through natural language and have actions carried out on their behalf \citep{yujia-tool-survey,agent-survey-renda,agent-survey-fdu,long-horizon-survey,multi-agent-survey}.
Through tool interfaces, agents can access external services and coordinate multi-step operations to fulfill these goals \citep{toolllm,toolorchestra,autoact,agent-world,agentscaler,toolrl}.
This opens the prospect of a unified service interface: users specify what they need, while agents take responsibility for finding and orchestrating the capabilities required to deliver it.

\begin{wrapfigure}{r}{0.5\textwidth}
    \vskip -0.25in
    \centering
    \includegraphics[width=0.5\textwidth]{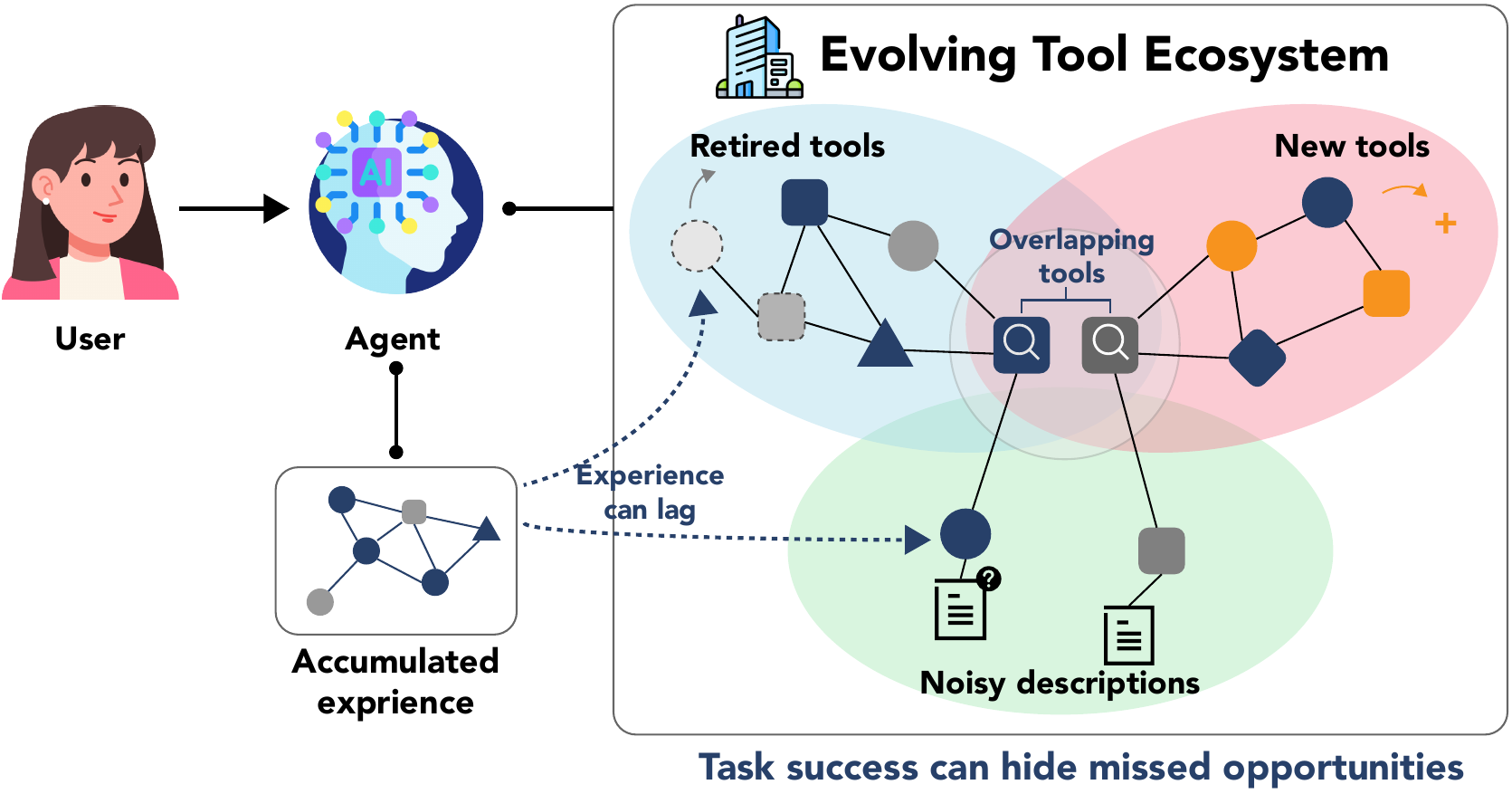}
    \vspace{-0.65cm}
    \caption{\small \textbf{Agent adaptation in evolving tool ecosystems.} Tool updates can invalidate learned workflows or introduce better alternatives, while overlapping tools and noisy descriptions hinder discovery.}
    \label{fig:intro}
    \vspace{-0.4cm}
\end{wrapfigure}
As shown in Fig.~\ref{fig:intro}, imagine a large company deploying an agent as a unified entry point to its services.
Behind this single interface lies a large tool ecosystem, with individual departments independently maintaining and updating their own tools.
Overlapping functionality and noisy or inaccurate descriptions make this catalog difficult to navigate, while ongoing updates continually reshape the available paths through it.
As departments retire or replace tools, established workflows may break, forcing the agent to search the broader ecosystem for alternatives.
Meanwhile, newly added tools may enable more efficient workflows, but these opportunities can remain hidden among the many services while familiar procedures continue to succeed.
Keeping pace therefore requires the agent to actively discover how the company's changing capabilities can be used and revise its procedural knowledge accordingly.
Otherwise, successful task execution can conceal a growing gap between what the ecosystem offers and what the agent has learned to accomplish with it.

Evaluating this challenge requires comparing an agent's behavior with what the current ecosystem makes possible for the same task.
Existing benchmarks assess large-scale tool retrieval and stateful execution \citep{toolret,meta-tool,appworld,taubench,toolathlon}, cost-aware replanning under changing execution conditions \citep{costbench,planbench,travelplanner,acpbench}, and robustness to tool evolution \citep{mcpevol-bench,toolqa-d}.
However, these settings largely examine these abilities separately, generally keeping service structures fixed or constraining tool discovery when services change.
\textit{They leave open whether agents can proactively discover efficient tool-use paths and promptly update their procedural knowledge as a large ecosystem evolves.}
Addressing this question calls for an evaluation of sustained agent competence against the moving capability frontier of an evolving tool ecosystem.

To address this gap, we introduce \textbf{\bench}, an executable benchmark spanning the full tool-use lifecycle in large, evolving ecosystems.
It comprises stateful services across 42 domains and 1,588 validated tasks requiring multi-turn user interaction.
Starting from these services, we generate six successive ecosystem states through capability-preserving transformations that refine, split, or integrate tool interfaces.
Older and newer tools may coexist or replace one another, changing how tasks can be completed while preserving access to the underlying capabilities.
For each state, we identify the most efficient task paths within the constructed solution space and use them as references for evaluating agent behavior.
By assessing tool discovery alongside stateful execution against these evolving references, {\bench} measures whether agents recover from disrupted workflows and exploit efficiency gains that successful execution alone may fail to reveal.

Evaluation shows that current agents struggle to sustain reliable and efficient execution in evolving tool ecosystems, with even the strongest evaluated model or harness achieving only 42.3\% task success.
Existing skill-learning methods provide initial gains, but these do not consistently translate into better adaptation as the ecosystem evolves, and skill maintenance can also incur substantial overhead.
These findings highlight the need to manage accumulated knowledge together with the changing tools through which it is applied.
We therefore introduce \textbf{\ours}, an ecosystem manager that coordinates the full lifecycle of tools and skills as a coupled resource system, from knowledge updates to downstream use.
By tracking skill--tool dependencies, {\ours} incrementally maintains the skill library as tools change, preserving reusable experience.
During execution, it jointly supplies relevant tools and procedural guidance, adapting this resource selection as the task unfolds.
This connects what the agent learns and retains with what it can effectively use in the current ecosystem.

Experiments show that {\ours} sustains stronger performance across ecosystem changes while reducing skill-maintenance overhead.
Its benefits extend beyond keeping existing workflows functional to exploiting more efficient solutions as they become available.
Further analyses show that preserving and selectively updating accumulated knowledge is more effective than repeatedly rebuilding it, while jointly providing tools and skills yields benefits that neither resource provides alone.



\section{Task Formulation}
We model agentic interaction as a partially observable sequential decision process $\mathcal{M}=(\mathcal{U},\mathcal{S},\mathcal{A},\mathcal{O},\mathcal{P},\mathcal{R})$.
Here, $\mathcal{U}$ is the space of user intents; each task typically has an intent $u\in\mathcal{U}$ not directly observable to the agent.
The state space $\mathcal{S}=\mathcal{S}_{E}\times\mathcal{S}_{H}$ combines external environment states $s^E$ and agent--user interaction states $s^H$, with $s=(s^E,s^H)$.
The action space $\mathcal{A}=\mathcal{A}_{\textrm{tool}}\cup\mathcal{A}_{\textrm{resp}}$ comprises executable tool calls $a=(f,\mathbf{x})\in\mathcal{A}_{\textrm{tool}}$, with tool $f$ and arguments $\mathbf{x}$, and natural-language messages $a=y\in\mathcal{A}_{\textrm{resp}}$ to the user.
The observation space $\mathcal{O}=\mathcal{O}_{\textrm{tool}}\cup\mathcal{O}_{\textrm{user}}$ comprises tool returns or errors and the user's initial request and subsequent replies, respectively.
With observable history $h_t=(o_0,a_0,\ldots,a_{t-1},o_t)$, action selection and environment transitions follow:
\begin{align}
a_t\sim\pi_\theta(\cdot\mid h_t),
\qquad
(s_{t+1},o_{t+1})\sim\mathcal{P}(\cdot\mid s_t,a_t,u).
\end{align}
Along $\tau=(o_0,a_0,o_1,a_1,\ldots,o_T,a_T)$, tool calls may modify $s_t^E$ and dialogue may modify $s_t^H$.
At completion, reward $\mathcal{R}(\tau,u)$ scores the final environment state and satisfaction of user constraints.

Each episode has an exogenous tool-ecosystem version $v\in\mathcal{V}$, fixed throughout execution, with action space $\mathcal{A}^{(v)}=\mathcal{A}_{\textrm{tool}}^{(v)}\cup\mathcal{A}_{\textrm{resp}}$, where $\mathcal{A}_{\textrm{tool}}^{(v)}$ contains all tool calls executable at $v$.
Across versions, changes to available functions and admissible arguments can alter feasible functions and successful action paths for the same intent $u$.
The policy must consider $h_t$ and $\mathcal{A}^{(v)}$ at each step $t$.

\begin{figure*}
    \centering
    \vspace{-1.cm}
    \includegraphics[width=1.\textwidth]{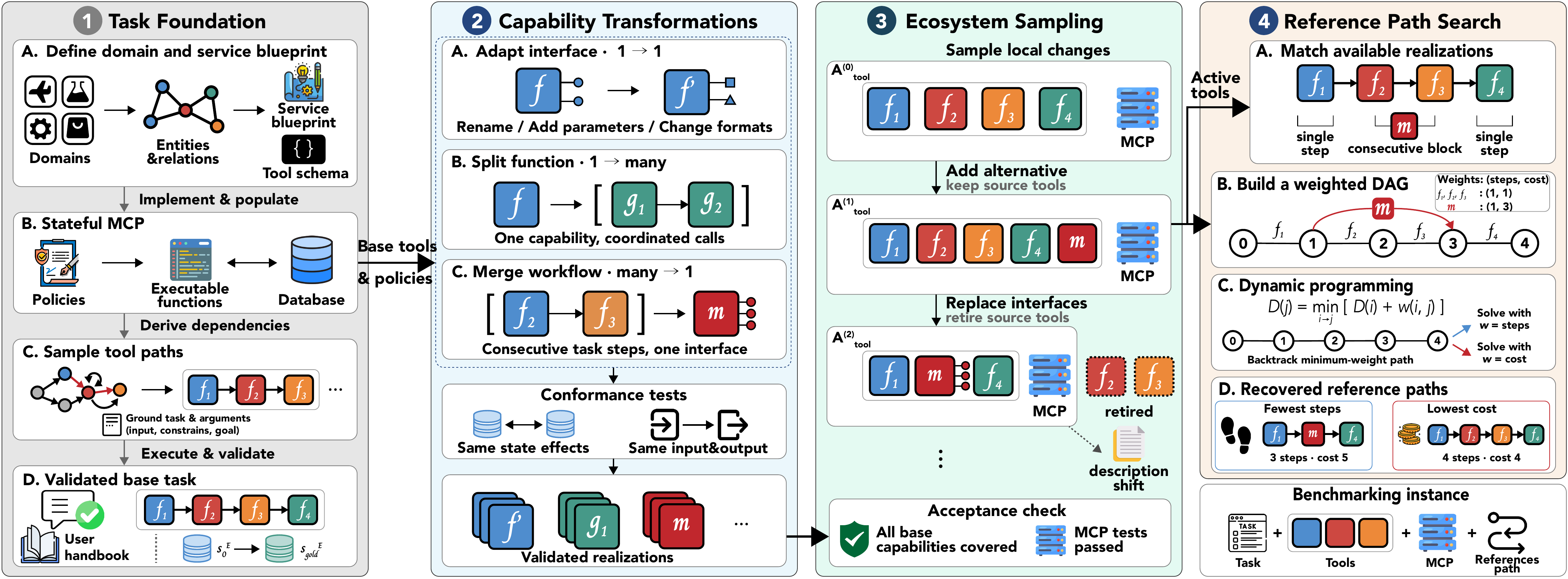}
    \vspace{-0.2cm}
    \caption{\small \textbf{Overview of {\bench} Construction.} We construct stateful environments and execution-validated tasks, then sample evolving tool ecosystems through validated capability-preserving transformations.
    Dynamic programming identifies step- and cost-minimal reference paths within the constructed realization space.}
    \label{fig:benchmark}
    \vspace{-0.4cm}
\end{figure*}

\section{{\bench}}

We construct {\bench} (Fig.~\ref{fig:benchmark}), an executable benchmark that systematically evaluates agent behavior in large and evolving tool ecosystems.
It combines stateful multi-turn tasks with composable, capability-preserving transformations, enabling scalable ecosystem evolution without sacrificing executability and cross-ecosystem compatibility.
{\bench} contains 1,588 tasks spanning 42 domains, with 1,191 tasks in the training set and 397 in the test set.
Across six ecosystem states, this yields 9,528 task–ecosystem instances.
Due to page limits, the detailed benchmark construction process and the statistics of {\bench} are described in Appx.~\ref{app:bench-con-detail}.

\subsection{Executable Environment and Task Construction}
\label{sec:exe-env}

\paragraph{Domain-Grounded Stateful Environments.}
Following \cite{agentskiller}, we expand seed domains into specialized, role-specific scenarios grouped by related activities.
For each group, we construct entity schemas and an entity–relationship graph around a primary user role.
These define coherent service blueprints specifying users, maintained information, and functions.
Validated functions are deterministically converted into interfaces with natural-language descriptions and JSON Schemas, forming the tool-call action space $\mathcal{A}_{\textrm{tool}}$.
We instantiate entity records and relationships in a mock database, validating schema and relational constraints to obtain the initial external state $s_0^E$.
MCP services implement these interfaces under policies governing parameter semantics, invocation constraints, state changes, and dependencies, returning results or errors in declared formats.

\vspace{-0.1cm}
\paragraph{Grounded Task Synthesis and Execution Validation.}
From the interfaces and policies, we derive a tool-dependency graph and sample paths under dependency and parameter-flow constraints.
Sampling favors state-querying starts while allowing state-changing starts when their preconditions hold.
To increase task difficulty, we join paths sharing a function and repeat intermediate subpaths, retaining only sequences with coherent function ordering, parameter flow, and task objectives.
Conditioned on the service blueprints, interfaces, policies, and database summaries, we generate natural-language task templates and user intents requiring every sampled function for a single coherent objective.
An independent evaluator screens templates for motivation clarity, coherence, function coverage, naturalness, and specificity.
For retained templates, we ground arguments in user-known information or preceding tool outputs and replay instantiated sequences from $s_0^E$ against the MCP service.
Only instances where every action succeeds with the intended observations and state transitions are retained, yielding reference trajectories $\tau_{\textrm{gold}}$ and terminal external states $s_{\textrm{gold}}^E$.

\vspace{-0.1cm}
\paragraph{Task-Conditioned User Simulation.}
An independent LLM-based user agent follows a general protocol and a private handbook specifying the complete intent $u$, motivation, grounded user information, and disclosure rules.
We generate an initial request $o_0$ that expresses the user's immediate need while leaving some task details for subsequent dialogue.
When the evaluated agent asks for missing task information, the user agent provides the corresponding details from the handbook in accordance with the task-specific disclosure rules.
Throughout the dialogue, the user agent speaks from the user's perspective, progressively disclosing only grounded information without mentioning functions or execution plans.
Responses are conditioned on the protocol, handbook, and visible dialogue, which contains only natural-language messages exchanged between the agents.
Interaction ends when the user agent indicates that its intent is satisfied or the request is outside the service scope.

\subsection{Evolving Tool Ecosystems Construction}
\label{sec:evol-con}
\paragraph{Composable Capability Transformations.}
We construct each domain's space of alternative capability realizations from its executable base tool set $\mathcal{A}_{\textrm{tool}}^{(0)}=\mathcal{A}_{\textrm{tool}}$.
A transformation $\phi:P\mapsto R$ maps a source block $P$ to a target realization $R$, each containing one or more functions from the same domain, while preserving the source capabilities under a different interface structure.
Function counts define three forms of transformation:
1) \textit{one-to-one}, renaming functions or parameters, adding parameters, or changing formats;
2) \textit{one-to-many}, decomposing a function into coordinated interfaces; and
3) \textit{many-to-one}, integrating functions into a unified interface.
Integration requires $P$ to be a contiguous subsequence of a validated same-domain task trajectory, grounding $R$ in a coherent, executable workflow.
Given $P$, an LLM conditioned on the domain's service blueprint, function schemas, and operational policies proposes $R$ and its executable implementation.
Each candidate records explicit lineage to its source functions and is instantiated in the corresponding stateful environment for automated validation.
Only transformations preserving required source capabilities, task-relevant state changes, and externally consumed outputs under equivalent inputs are retained.
Local transformations incrementally expand this space without reconstructing domains or services.

\vspace{-0.1cm}
\paragraph{Capability-Preserving Ecosystem Sampling.}
We organize validated realizations by source-to-target lineage and sample each version's active interface set $\mathcal{A}_{\textrm{tool}}^{(v)}$.
Starting from all executable base interfaces in $\mathcal{A}_{\textrm{tool}}^{(0)}$, subsequent versions apply a bounded number of validated transformations to $\mathcal{A}_{\textrm{tool}}^{(v-1)}$, leaving other interfaces unchanged.
For gradual updates and backward compatibility, we introduce $R$ and either retain $P$, requiring selection among overlapping interfaces, or deactivate $P$, requiring agents to access and compose capabilities through $R$.
We recursively verify coverage through lineage, rejecting transitions unless every base capability remains accessible through active interfaces, directly or through coordinated realizations.
For accepted states, we adapt the domain's operational policies and assemble validated implementations into executable MCP services.
We then optionally perturb selected function descriptions to simulate incomplete, compressed, or stale documentation, preserving available functions, parameter schemas, and execution semantics.

\vspace{-0.1cm}
\paragraph{Task-Level Optimal Realization Search.}
Across versions, user intent and target state remain fixed, while task realizations may change.
We extract each task's ordered base-function sequence $(f_1,\ldots,f_n)$ from its validated trajectory $\tau_{\textrm{gold}}$.
At version $v$, a task realization comprises active realizations whose source blocks exactly cover this sequence in order.
The \textbf{reference path} minimizes the total value of a specified efficiency criterion over feasible task realizations within the constructed realization space.
Step counts tool calls, while cost sums per-call scores (1--10) assigned by an LLM based on each tool's definition and implementation to reflect its potential execution cost.
Let $\Phi^{(v)}(P)$ denote the realizations of $P$ available in $\mathcal{A}_{\textrm{tool}}^{(v)}$.
Partitioning the sequence into contiguous source blocks and selecting one active realization per block gives:
\vspace{-0.3cm}
\begin{align}
\rho_i^{(v,c)}
=
\underset{\substack{
P_1\oplus\cdots\oplus P_m=(f_1,\ldots,f_n)\\
R_j\in\Phi^{(v)}(P_j)
}}{\arg\min}
\sum_{j=1}^{m}c^{(v)}(R_j),
\end{align}
where $\oplus$ denotes sequence concatenation and $c^{(v)}(R_i)$ is the criterion value of $R_i$.
Dynamic programming over active realizations in the task's domain separately minimizes function-call count and execution cost, yielding step-minimal and cost-minimal reference paths for each task and version.
We replay each reference path in its corresponding MCP environment and retain only paths whose calls succeed and reproduce the required state changes.

\subsection{Evaluation Protocol and Metrics}
\label{sec:metrics}
\textbf{Outcome-based Task Success Rate.}
For task $i$ at version $v$, the agent starts from $s_{0,i}^E$, and yields trajectory $\tau_i^{(v)}$ with terminal state $s_{T,i}^{E,(v)}$.
Let $\Delta(s,s')$ denote database changes between external states.
The required and agent-produced changes are:
\begin{align}
\Delta_i^{\mathrm{gold}}
=
\Delta(s_{0,i}^E,s_{\mathrm{gold},i}^E),
\qquad
\Delta_i^{(v)}
=
\Delta(s_{0,i}^E,s_{T,i}^{E,(v)}).
\end{align}
A task succeeds when agent's changes cover all changes required by the validated gold execution:
\begin{align}
S_i^{(v)}
=
\mathcal{R}(\tau_i^{(v)},u_i)
=
\mathbb{I}\!\left[
\Delta_i^{\mathrm{gold}}
\subseteq
\Delta_i^{(v)}
\right].
\end{align}
For $N_v$ tasks evaluated at version $v$, the success rate $
\mathrm{SR}^{(v)}
=
\frac{1}{N_v}
\sum_{i=1}^{N_v}S_i^{(v)}.
$

\textbf{Tool Retrieval Quality.}
For task $i$ at version $v$, let $\widehat{\mathcal A}_i^{(v)}$ and $\mathcal A_i^{(v)}$ denote the final tool set exposed to the agent after retrieval and the reference-path tools, respectively, both subsets of $\mathcal A_{\textrm{tool}}^{(v)}$.
Task-level retrieval precision, recall, and F1 are:
\begin{align}
\mathrm{P}_i^{(v)}
=
\tfrac{
|\widehat{\mathcal A}_i^{(v)}\cap\mathcal A_i^{(v)}|
}{
|\widehat{\mathcal A}_i^{(v)}|
},
\quad
\mathrm{R}_i^{(v)}
=
\tfrac{
|\widehat{\mathcal A}_i^{(v)}\cap\mathcal A_i^{(v)}|
}{
|\mathcal A_i^{(v)}|
},
\quad
\mathrm{F1}_i^{(v)}
=
\tfrac{
2\mathrm{P}_i^{(v)}\mathrm{R}_i^{(v)}
}{
\mathrm{P}_i^{(v)}+\mathrm{R}_i^{(v)}
}.
\end{align}

\textbf{Success-weighted Path Efficiency.}
Success-weighted Path Efficiency (SPE) measures execution efficiency relative to reference paths.
For task $i$ at version $v$, let $\mathcal{W}_i^{(v)}$ and $\widehat{\mathcal{W}}_i^{(v)}$ denote the reference-path and agent tool-call trajectory weights under the same criterion.
We define
\begin{align}
\mathrm{SPE}^{(v)}
=
\tfrac{1}{N_v}
\sum_{i=1}^{N_v}
S_i^{(v)}
\min\left(
\tfrac{\mathcal{W}_i^{(v)}}
{\widehat{\mathcal{W}}_i^{(v)}},
1
\right).
\end{align}
Using function-call count and execution cost yields Step-SPE and Cost-SPE, respectively.

\textbf{Improvement Discovery and Degradation Adaptation.}
To evaluate frontier adaptation, we identify tasks whose reference-path weights decrease or increase across adjacent ecosystem versions:
\begin{align}
\mathcal{D}_{\downarrow}^{(v)}
=
\left\{
i \mid
\mathcal{W}_{i}^{(v)}
<
\mathcal{W}_{i}^{(v-1)}
\right\},
\qquad
\mathcal{D}_{\uparrow}^{(v)}
=
\left\{
i \mid
\mathcal{W}_{i}^{(v)}
>
\mathcal{W}_{i}^{(v-1)}
\right\}.
\end{align}
The Improvement Discovery Rate (IDR) and Degradation Adaptation Rate (DAR) are:
\begin{align}
\small
\mathrm{IDR}^{(v)}
=
\tfrac{1}{\left|\mathcal{D}_{\downarrow}^{(v)}\right|}
\sum_{i\in\mathcal{D}_{\downarrow}^{(v)}}
S_i^{(v)}
\min\left(
\tfrac{\mathcal{W}_{i}^{(v)}}
{\widehat{\mathcal{W}}_{i}^{(v)}},
1
\right),
\mathrm{DAR}^{(v)}
=
\tfrac{1}{\left|\mathcal{D}_{\uparrow}^{(v)}\right|}
\sum_{i\in\mathcal{D}_{\uparrow}^{(v)}}
S_i^{(v)}
\min\left(
\tfrac{\mathcal{W}_{i}^{(v)}}
{\widehat{\mathcal{W}}_{i}^{(v)}},
1
\right).
\end{align}
IDR measures the discovery of newly shortened or cheaper paths.
DAR measures effectiveness when optimal paths become longer or costlier.
Steps and cost yield Step-IDR/DAR and Cost-IDR/DAR.

\begin{table}[t]
    \centering
    \renewcommand\arraystretch{1.}
    \vspace{-1.cm}
    \caption{\small \textbf{Model evaluation on {\bench}.} The best results for each metric are highlighted in \textbf{bold}. Current models struggle to reach the capability frontier, with limited task success and inefficient execution.}
    \label{tab:evaluation_models}
    \vskip -0.2cm
    \scalebox{.95}{
    \begin{tabular}{l|c|ccc|cc|cc}
        \toprule
        \multirow{2}{*}{\textbf{Model}}
        & \multirow{2}{*}{\textbf{SR}}
        & \multicolumn{3}{c|}{\textbf{Tool Retrieval}}
        & \multicolumn{2}{c|}{\textbf{SPE}}
        & \multicolumn{2}{c}{\textbf{SPE/SR}} \\
        \cmidrule{3-5}
        \cmidrule{6-7}
        \cmidrule{8-9}
        &
        & \textbf{P}
        & \textbf{R}
        & \textbf{F1}
        & \textbf{Step}
        & \textbf{Cost}
        & \textbf{Step}
        & \textbf{Cost} \\
        \midrule

        \multicolumn{9}{c}{\textbf{\textit{Proprietary Models}}} \\
        \midrule
        Codex-GPT-5.5
        & \textbf{42.27}
        & 19.90 & \textbf{90.36} & \textbf{32.04}
        & \textbf{26.93} & \textbf{27.90}
        & 63.71 & 66.00 \\

        Gemini-3.6-Flash
        & 41.65
        & 19.30 & 87.51 & 31.13
        & 25.49 & 25.88
        & 61.20 & 62.14 \\

        GPT-5.5
        & 31.74
        & \textbf{20.21} & 81.08 & 31.86
        & 23.11 & 23.42
        & \textbf{72.81} & \textbf{73.79} \\

        Claude-Sonnet-5
        & 29.97
        & 19.36 & 76.93 & 30.41
        & 20.48 & 20.60
        & 68.34 & 68.74 \\

        \midrule
        \multicolumn{9}{c}{\textbf{\textit{Open-source Models}}} \\
        \midrule
        DeepSeek-V4-Pro
        & 33.29
        & 18.65 & 79.00 & 29.65
        & 13.49 & 14.31
        & 40.52 & 42.99 \\

        GLM-5.2
        & 32.28
        & 19.32 & 78.25 & 30.47
        & 19.39 & 19.84
        & 60.07 & 61.46 \\

        Qwen-3.8-Max
        & 30.10
        & 18.46 & 80.02 & 29.49
        & 15.84 & 16.84
        & 52.62 & 55.94 \\

        Kimi-K3
        & 28.30
        & 19.20 & 77.01 & 30.20
        & 17.78 & 17.83
        & 62.83 & 63.00 \\

        GPT-oss-120B
        & 27.00
        & 18.24 & 69.98 & 28.40
        & 16.87 & 17.16
        & 62.48 & 63.56 \\



        \bottomrule
    \end{tabular}
    }
\end{table}

\section{Evaluating Agents in Evolving Tool Ecosystems}

\paragraph{Evaluation Setup.}
\label{sec:eval-setup}
We evaluate whether current powerful models can reach the capability frontier of each of six tool-ecosystem versions and whether automated skill methods can track this frontier by accumulating experiences as the ecosystem evolves.
For models that expose a configurable reasoning effort, we set it to \texttt{high} and retain the recommended defaults for all other inference parameters.
The user model is fixed to DeepSeek-V3.2 throughout all experiments.
All controlled model ablations and all comparisons in the skill-method study use DeepSeek-V3.2 as the task model.

\subsection{Models Struggle to Reach the Current Frontier}
We examine whether current models can independently reach the capability frontier of each tool-ecosystem version.
As shown in Tab.~\ref{tab:evaluation_models}, even the strongest model (Gemini-3.6-Flash) and harness (Codex-GPT-5.5) achieve SRs of only 41.65\% and 42.27\%, respectively, while neither exceeds 30\% in Step- or Cost-SPE.
Tool retrieval exhibits a pronounced imbalance: Codex-GPT-5.5 retrieves 90.36\% of the relevant tools, yet attains only 19.90\% precision and an F1 score of 32.04\%, a pattern that persists across models, indicating that models often include relevant tools but struggle to exclude irrelevant candidates from the broader ecosystem.
Moreover, success does not necessarily imply efficient execution.
For example, although DeepSeek-V4-Pro slightly outperforms GLM-5.2 in SR, it trails in both Step- and Cost-SPE, and Step- and Cost-SPE/SR.

\begin{wrapfigure}{r}{0.45\textwidth}
    \vskip -0.3in
    \centering
    \includegraphics[width=0.45\textwidth]{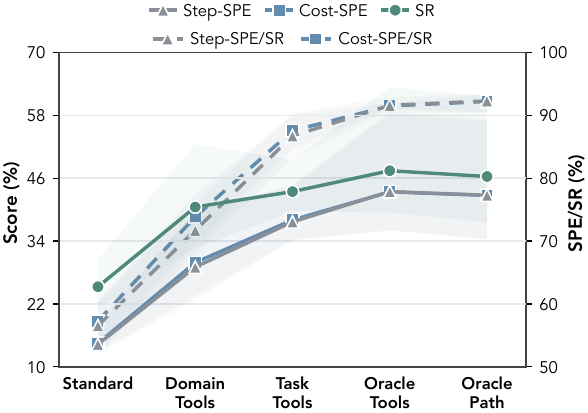}
    \vspace{-0.5cm}
    \caption{\small \textbf{Ablations on Tool Provision}.}
    \label{fig:evaluation_ablation}
    \vspace{-0.4cm}
\end{wrapfigure}
We next conduct a controlled diagnostic to probe barriers to identifying and executing efficient tool-use paths.
Specifically, we compare \textit{Standard} with four increasingly informative conditions.
\textit{Domain Tools} restricts retrieval to the task domain, whereas \textit{Task Tools} provides all tools used by the task across versions to test whether models can distinguish currently available and efficient options from unavailable or overlapping alternatives.
\textit{Oracle Tools} retains only relevant, current-version tools, while \textit{Oracle Path} additionally provides the reference tool path to isolate path-planning difficulty.
As shown in Fig.~\ref{fig:evaluation_ablation}, SR and both SPE variants rise steadily from \textit{Standard} through \textit{Domain Tools} and \textit{Task Tools} to \textit{Oracle Tools}, whereas \textit{Oracle Path} yields no further improvement.
The gains from \textit{Standard} to \textit{Domain Tools} and \textit{Task Tools} show that removing domain- and task-irrelevant candidates eases tool identification.
More importantly, the pronounced SPE/SR gain from \textit{Task Tools} to \textit{Oracle Tools} shows that models struggle to distinguish efficient, version-valid options from unavailable or overlapping alternatives.
The negligible change under \textit{Oracle Path} instead indicates that the remaining gap lies less in high-level planning than in correctly and efficiently using available tools.
Overall, current models struggle both to retrieve relevant tools precisely and to execute their efficient realizations in evolving ecosystems.

\begin{figure*}
    \centering
    \vspace{-1.cm}
    \includegraphics[width=1.\textwidth]{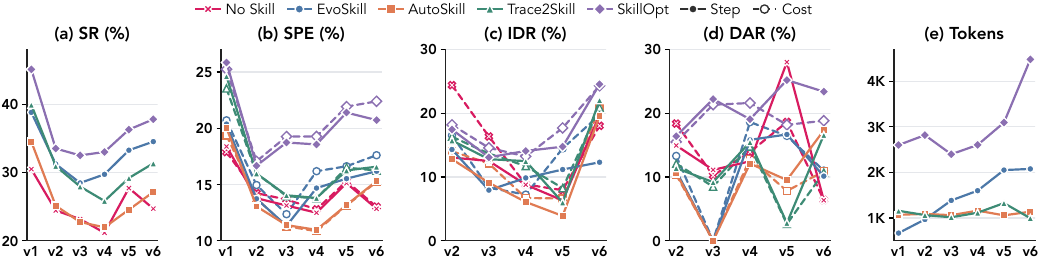}
    \vspace{-0.5cm}
    \caption{\small   \textbf{Evaluation of automated skill methods across ecosystem versions.}
    Current skill evolution strategy can provide initial performance gains but cannot consistently adapt to the moving capability frontier.}
    \label{fig:evaluation_skill}
    \vspace{-0.4cm}
\end{figure*}

\subsection{Skill Methods Struggle to Track the Moving Frontier}
Our preceding diagnosis shows that models struggle primarily to identify and efficiently execute currently valid tool-use paths in an evolving ecosystem.
So, we examine whether automated skill generation and optimization methods can overcome this limitation by accumulating experience.
As shown in Fig.~\ref{fig:evaluation_skill}, we compare four automated skill methods: EvoSkill \citep{evoskill}, AutoSkill \citep{autoskill}, Trace2Skill \citep{trace2skill}, SkillOpt \citep{skillopt}, against the \textit{No Skill} baseline across six ecosystem versions (Please refer to Appx.~\ref{app:baselines} for baseline details).
At \textit{v1}, all four methods outperform \textit{No Skill} in SR and both SPE variants, showing that accumulated experience provides useful initial guidance.
However, these gains do not persist uniformly: AutoSkill falls below \textit{No Skill} in SR at \textit{v3} and \textit{v5}, while AutoSkill and EvoSkill also underperform it in SPE at several intermediate versions.
This limitation is more pronounced in DAR: at \textit{v3}, both EvoSkill and AutoSkill drop to 0 in Step- and Cost-DAR, whereas \textit{No Skill} retains 11.11\% and 10.25\%, respectively.
Moreover, \textit{No Skill} achieves the highest Cost-IDR at \textit{v2} and both DAR variants at \textit{v5}, indicating that accumulated skills do not consistently improve adaptation to ecosystem changes.
Accordingly, although SkillOpt leads in SR and SPE across versions, its advantage does not extend consistently to IDR and DAR, showing that stronger current-version performance does not necessarily imply better frontier tracking.
This stronger current-version performance also incurs substantially higher maintenance costs, with SkillOpt averaging about 3,000 K/task for skill generation and optimization, roughly two to three times the other methods.
These findings show that generating and optimizing skills in isolation is insufficient for sustained and efficient frontier tracking.
Reliable adaptation requires tools and skills to be jointly maintained and provisioned as a unified, evolving ecosystem.

\begin{figure*}
    \centering
    \vspace{-1.cm}
    \includegraphics[width=.92\textwidth]{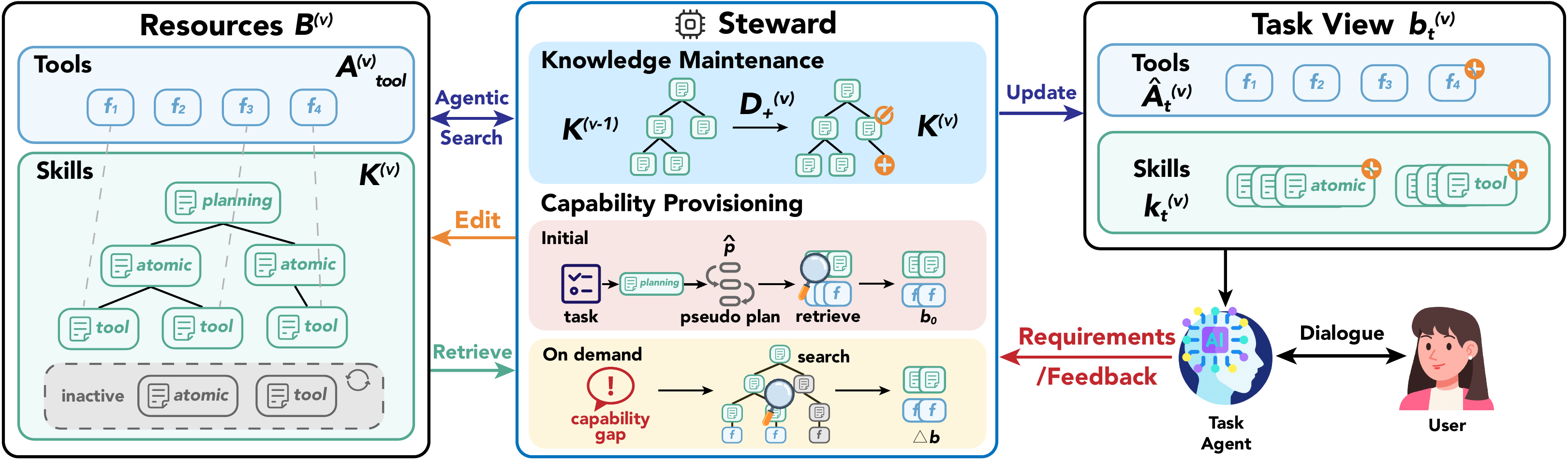}
    \vspace{-0.2cm}
    \caption{\small \textbf{Overview of {\ours}.} As an ecosystem manager, {\ours} maintains shared tool and skill resources, updating hierarchical skills using verified experience and current tool availability.
    It initializes the task agent's view through planning and retrieval, then dynamically updates it as capability gaps emerge.}
    \label{fig:method}
    \vspace{-0.4cm}
\end{figure*}

\section{\ours: Managing Tools and Skills in Evolving Ecosystems}
\label{sec:steward}
\textbf{The Ecosystem Manager.}
We introduce {\ours}, a model-driven ecosystem manager that coordinates the maintenance of tool and skill resources and their provision to agents for task execution.
At version $v$, {\ours} maintains a shared resource state $\mathcal B^{(v)}=(\mathcal A_{\textrm{tool}}^{(v)},\mathcal K^{(v)})$, where $\mathcal K^{(v)}$ is a persistent skill library that records procedural knowledge, tool dependencies, and availability status.
Through dedicated management interfaces, {\ours} updates skills, retrieves relevant tools and knowledge, and inspects tool schemas to configure the resources available to the task agent.
Conditioned on the interaction history $h_t$, {\ours} constructs a task-specific view $b_t^{(v)}=(\widehat{\mathcal A}_t^{(v)},k_t^{(v)})$ of these resources, comprising the exposed tools $\widehat{\mathcal A}_t^{(v)}\subseteq\mathcal A_{\textrm{tool}}^{(v)}$ and the accompanying procedural context $k_t^{(v)}$.
The task agent then acts according to $a_t\sim\pi_\theta(\cdot\mid h_t,b_t^{(v)})$, while {\ours} maintains $\mathcal K^{(v)}$ across versions and adapts $b_t^{(v)}$ to evolving task requirements.
Due to page limits, please refer to Appx.~\ref{app:method-detial} for more method details and Appx.~\ref{app:prompts} for prompts used in {\ours}.

\textbf{Knowledge Maintenance.}
We initialize skills at $v=1$ by following \cite{skillx}.
As tools evolve, {\ours} incrementally updates $\mathcal K^{(v-1)}$ using the current tool catalog and newly verified execution experience while preserving reusable knowledge.
The library follows a three-level hierarchy: \textit{planning skills} capture task-level workflows, \textit{atomic skills} encode the reusable operations within these workflows, and \textit{tool skills} specify the invocation requirements of individual tools and are linked to the atomic skills that use them.
At version $v$, the manager takes as input $\mathcal K^{(v-1)}$, $\mathcal A_{\textrm{tool}}^{(v)}$, and a set $\mathcal D_{+}^{(v)}$ of successful training trajectories derived from the task agent and verified through environment-state evaluation.
Conditioned on $\mathcal K^{(v-1)}$, the manager distills $\mathcal D_{+}^{(v)}$ into local edits to the existing skill hierarchy, adding or revising skills at the appropriate levels and updating their links.
The manager then marks skills as active or inactive according to the availability of their tool dependencies in $\mathcal A_{\textrm{tool}}^{(v)}$, retaining inactive skills for reactivation when their dependencies become available again.
The resulting library $\mathcal K^{(v)}$ preserves both partitions, with retrieval indices and the skill graph rebuilt over its active subset $\mathcal K_{\textrm{act}}^{(v)}$ for capability provisioning.

\textbf{Capability Provisioning.}
For each task, {\ours} jointly selects tools from $\mathcal A_{\textrm{tool}}^{(v)}$ and procedural knowledge from $\mathcal K_{\textrm{act}}^{(v)}$ to construct $b_t^{(v)}$ according to the task's current requirements.
At the start of an episode, the manager retrieves planning skills relevant to the initial user request and uses them to construct a task-specific provisional plan $\hat p$.
For each step of $\hat p$, the manager retrieves atomic skills and combines their referenced tools with directly retrieved tools.
Selected atomic skills and their linked tool skills form $k_0^{(v)}$, while candidate tools available in the current version form $\widehat{\mathcal A}_0^{(v)}$, together constituting $b_0^{(v)}$.
When new user information or execution feedback reveals capability gaps, the manager supplements the task view through lightweight retrieval or agentic search and inspection across $\mathcal A_{\textrm{tool}}^{(v)}$ and $\mathcal K_{\textrm{act}}^{(v)}$.
The manager adds verified tools to $\widehat{\mathcal A}_t^{(v)}$ and updates $k_t^{(v)}$ with procedural guidance and, when needed, the full text of selected skills.
By coupling incremental knowledge maintenance with on-demand capability provisioning, {\ours} aims to reduce the costs of skill updates and task execution while preserving task success as the tool ecosystem evolves.

\begin{figure*}
    \centering
    \vspace{-1.cm}
    \includegraphics[width=1.\textwidth]{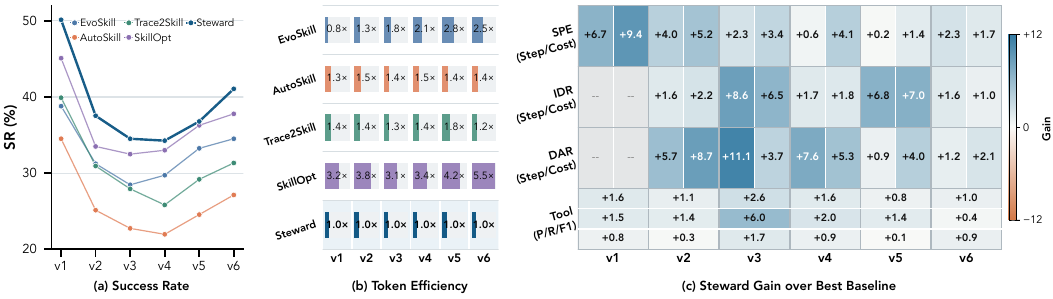}
    \vspace{-0.5cm}
    \caption{\small \textbf{Evaluation of {\ours} in evolving tool ecosystems.} \textbf{(a) Task success.} {\ours} achieves the highest SR across all six ecosystem versions.
    \textbf{(b) Relative maintenance cost.} Baselines require 1.2--5.5$\times$ as many maintenance tokens as {\ours} from $v_2$ onward.
    \textbf{(c) Percentage-point gains over the best baseline per metric and version.} {\ours} consistently improves execution efficiency, frontier tracking, and tool retrieval.}
    \label{fig:exp_main}
    \vspace{-0.4cm}
\end{figure*}

\section{Experimental Results for {\ours}}
\label{sec:exp-results}
\paragraph{{\ours} Tracks the moving frontier efficiently.}
We evaluate whether the joint management of tools and skills enables {\ours} to sustain task success and efficiently track the moving capability frontier using DeepSeek-V3.2.
As shown in Fig.~\ref{fig:exp_main}, we compare {\ours} with four automated skill-based methods across six tool-ecosystem versions under the same task model, user model, and evaluation protocol.
In Fig.~\ref{fig:exp_main}(a), {\ours} achieves the highest SR at every version, averaging 39.04\% versus 36.36\% for SkillOpt, showing that its advantage persists beyond initial skill construction as the ecosystem evolves.
Fig.~\ref{fig:exp_main}(b) further shows that {\ours} incurs the lowest maintenance cost from \textit{v2} onward, with the baselines consuming 1.2--5.5$\times$ as many skill-maintenance tokens.
Thus, the higher task success is achieved without increasing skill-update expenditure, distinguishing {\ours} from methods that rely on increasingly costly skill generation or optimization.
To examine whether this advantage extends beyond SR, Fig.~\ref{fig:exp_main}(c) reports the percentage-point gain of {\ours} over the strongest baseline for each version and metric.
Most notably, {\ours} improves Step-/Cost-IDR by 4.04\%/3.69\% and Step-/Cost-DAR by 5.28\%/4.76\% on average, showing that its advantage extends from current-version success to discovering improved paths and recovering from degraded ones across ecosystem changes.
These dynamic gains are accompanied by average Step-/Cost-SPE improvements of 2.68\%/4.21\% and consistent gains in tool precision, recall, and F1, indicating more efficient execution and more effective capability provisioning.
Together, these results establish that {\ours} combines stronger frontier tracking with lower maintenance overhead, motivating the following analyses of which management mechanisms produce these gains.

\begin{wraptable}{l}{0.7\textwidth}
\centering
\renewcommand\arraystretch{1.1}
\vspace{-0.65cm}
\caption{\small \textbf{Ablations on cross-version knowledge maintenance.}
Across both models, incremental maintenance leads on all performance metrics while requiring fewer maintenance tokens than full reconstruction.}
\vspace{-0.2cm}
\scalebox{.55}{
\begin{tabular}{l|l|ccccc|c}
\toprule
\multirow{2}{*}{\textbf{Model}}
& \multirow{2}{*}{\textbf{Strategy}}
& \multicolumn{5}{c|}{\textbf{Avg. Results (\%)}}
& \multirow{2}{*}{\shortstack{\textbf{Tokens}\\\textbf{(K/task)}}} \\
\cmidrule{3-7}
&
& \textbf{SR}
& \textbf{SPE}
& \textbf{IDR}
& \textbf{DAR}
& \textbf{Tool F1}
& \\
\midrule

\multirow[c]{3}{*}{\textbf{DeepSeek-V3.2}}
& Frozen Library
& 33.90
& 20.34/21.18
& 17.96/19.56
& 24.29/22.16
& 35.96
& \textbf{0.00} \\

& Full Reconstruction
& 34.61
& 21.06/21.95
& 17.62/18.80
& 24.24/21.37
& 33.56
& 1,271.10 \\

& \textbf{Incremental Maintenance}
& \textbf{36.83}
& \textbf{21.10/23.15}
& \textbf{20.83/21.29}
& \textbf{26.53/23.84}
& \textbf{39.37}
& 769.55 \\

\midrule

\multirow[c]{3}{*}{\textbf{GPT-5.5}}
& Frozen Library
& 39.60
& 28.23/29.26
& 25.65/28.11
& 28.69/26.17
& 27.20
& \textbf{0.00} \\

& Full Reconstruction
& 38.99
& 27.75/28.69
& 27.58/28.96
& 26.85/24.48
& 26.50
& 1,449.60 \\

& \textbf{Incremental Maintenance}
& \textbf{40.81}
& \textbf{30.67/31.62}
& \textbf{29.11/31.27}
& \textbf{31.23/29.62}
& \textbf{32.54}
& 1,009.02 \\

\bottomrule
\end{tabular}
}
\vskip -0.1in
\label{tab:ablation_maintenance}
\end{wraptable}
\paragraph{Incremental maintenance balances knowledge reuse and adaptation.}
To isolate cross-version maintenance, strategies within each model start from the same skill library at \textit{v1} and differ only in their maintenance from \textit{v2} onward.
As shown in Table~\ref{tab:ablation_maintenance}, incremental maintenance achieves the highest average SR for both DeepSeek-V3.2 and GPT-5.5 (36.83\% and 40.81\%), together with the best SPE, IDR, DAR, and tool F1.
Compared with full reconstruction, it improves SR by 2.22 and 1.82 percentage points while reducing maintenance tokens by 39.5\% and 30.4\%, respectively.
Freezing avoids update costs but yields lower SR than incremental maintenance; full reconstruction even underperforms freezing on GPT-5.5 (38.99\% vs.\ 39.60\%) despite its update cost.
These results suggest that targeted updates better balance the reuse of accumulated knowledge with adaptation to ecosystem changes.

\begin{wrapfigure}{r}{0.5\textwidth}
    \vskip -0.15in
    \centering
    \includegraphics[width=0.5\textwidth]{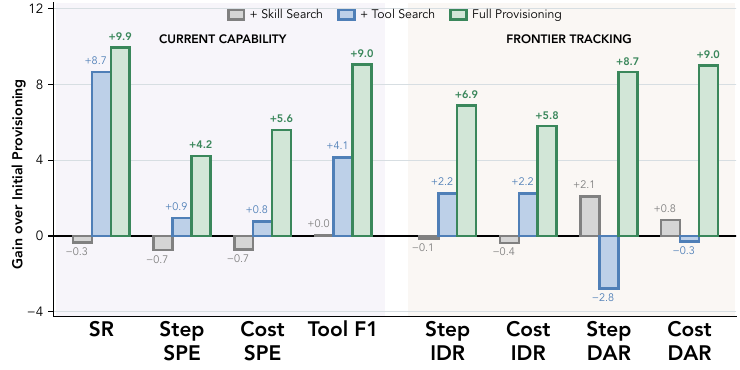}
    \vspace{-0.5cm}
    \caption{\small \textbf{Ablations on capability provisioning.}
    Combining dynamic tool and skill search yields the largest gains over initial provisioning across all metrics, demonstrating their complementary roles in task execution.}
    \label{fig:exp_ablation}
    \vspace{-0.2cm}
\end{wrapfigure}
\paragraph{Dynamic tool and skill search are complementary.}
We ablate capability provisioning on DeepSeek-V3.2 by taking initial task view construction as the baseline and adding dynamic skill search, tool search, or both.
As shown in Fig.~\ref{fig:exp_ablation}, skill search alone yields little improvement in SR, SPE, IDR, or Tool F1, indicating that procedural knowledge is difficult to operationalize when the corresponding tools are not dynamically exposed.
Tool search alone improves SR and Tool F1 by 8.65\% and 4.14\% percentage points, respectively, but provides limited SPE and IDR gains and
reduces Step-DAR by 2.76\%, showing that finding available tools does not ensure their efficient or adaptive use.
Combining both modules improves every metric; relative to tool search alone, \textit{Full Provisioning} raises Step-/Cost-DAR by 11.41\%/9.30\% percentage points and Tool F1 by 4.90\% points.
These results show that tool search exposes valid implementations while skill search supplies the procedural guidance needed to use them, making their joint management essential for capability provisioning.

\section{Related Work}

\paragraph{Benchmarks for Tool-Using Agents.}
As LLMs become autonomous agents, tool ecosystems mediate discovery and use of external capabilities, motivating benchmarks spanning a broader tool-use lifecycle \citep{tool-learning-survey,yujia-tool-survey}.
Early benchmarks assess API selection, argument generation, multi-tool composition, and planning \citep{apibench,toolllm,api-bank,toolalpaca,bfcl,t-eval,acebench}.
Stateful successors evaluate multi-turn interaction in environments with persistent tool effects, where success requires satisfying user constraints \citep{appworld,taubench,tau2bench,toolsandbox,toolathlon}.
Complementary benchmarks study retrieval from large, heterogeneous tool repositories \citep{iterative-tool-retrieval,toolret,meta-tool,toolgen,planbench-xl}, while efficiency-oriented benchmarks evaluate resource constraints, planning or operational costs, and replanning under changing costs, preferences, or tool availability \citep{planbench,saycanpay,acpbench,travelplanner,userbench,costbench}.
Recent work probes robustness through API-level perturbations or controlled mutations of MCP servers \citep{toolqa-d,mcpevol-bench}.
Yet these capabilities remain largely evaluated in isolation, leaving the end-to-end tool-use lifecycle in evolving ecosystems underexplored.
{\bench} addresses this gap as the first benchmark unifying large-scale tool retrieval, multi-turn stateful execution, and efficiency-aware adaptation within a controlled, evolving tool ecosystem.

\paragraph{Skill Learning and Evolution for LLM Agents.}
Agents should convert interaction experience into reusable procedural knowledge for continual improvement.
Early work distills trajectories into natural-language guidance or reusable workflows, while later methods represent skills as executable or compositional routines \citep{expel,autoguide,automanual,awm,wkm,voyager,learnact,skillweaver,cua-skill}.
Recent approaches automate skill construction and refinement through trajectory consolidation, optimization, reinforcement learning, verification, and persistent accumulation \citep{autoskill,trace2skill,evoskill,skillopt,skillprox,skillcommit,wikiskill}, with complementary work organizing growing libraries through hierarchies, dependencies, graphs, and lifecycle management \citep{skillx,skillsvote,skillops,skilldag,skillnet}.
Yet these methods largely evolve skills over a stable operational substrate and emphasize eventual task success, leaving open whether procedural knowledge can adapt quickly enough to sustain efficient execution as the capability frontier moves.
{\ours} recasts skill evolution as ecosystem-level adaptation, coupling tools with procedural knowledge to preserve reusable experience and track the evolving efficiency frontier.

\section{Conclusion}
We introduce {\bench}, an executable benchmark evaluating sustained agent competence against the moving capability frontier of large, evolving tool ecosystems.
The evaluated agents struggle to sustain reliable and efficient execution, while initial gains from accumulated skills do not consistently translate into better adaptation as tools change.
{\ours} couples incremental skill maintenance with dynamic provision of tools and procedural guidance, improving task performance and adaptation while reducing maintenance overhead.
These results support managing tools and procedural knowledge as interdependent resources, keeping accumulated experience useful as the ecosystem evolves.

\subsection*{AI use statement}


We used Large Language Models solely as writing and visualization assistants during the preparation of this paper, including for refining language, improving sentence structure, and designing the presentation of selected figures.
Any use of LLMs in our experiments was conducted exclusively for scientific research and has been transparently described in the Experimental Settings and Reproducibility Statement.
All LLM-assisted materials were carefully reviewed by the authors. We assume full responsibility for the final manuscript and all associated content, including its text, claims, analyses, and artifacts.


\subsection*{Ethics statement}

This study follows established ethical principles and responsible research practices. All data were obtained or synthesized from publicly accessible sources, with no proprietary, confidential, or otherwise restricted information involved. The original sources of the data are properly acknowledged and cited throughout the paper.
We expect users of our benchmark to employ it responsibly and in accordance with principles of fairness, transparency, and accountability. Applications that may cause harm or produce adverse societal consequences are not permitted.

\subsection*{Reproducibility statement}

The construction of executable environments and tasks is described in \S\ref{sec:exe-env} and Appx.~\ref{app:exe-env}, while ecosystem evolution and reference-path search are detailed in \S\ref{sec:evol-con} and Appx.~\ref{app:evol-con}.
Dataset statistics are reported in Appx.~\ref{app:statistics}, and the evaluation protocol and metrics are defined in \S\ref{sec:metrics}.
The design of {\ours}, including incremental knowledge maintenance and dynamic provisioning of tools and skills, is presented in \S\ref{sec:steward}.
The settings for baseline comparisons and controlled ablations are reported in \S\ref{sec:eval-setup}, \S\ref{sec:exp-results}, and Appx.~\ref{app:baselines}.
The prompts used in our study are provided in Appx.~\ref{app:prompts}.



\bibliography{iclr2027_conference}
\bibliographystyle{iclr2027_conference}

\appendix
\section{Details of {\bench} Construction}
\label{app:bench-con-detail}

\subsection{Executable Environment and Task Construction}
\label{app:exe-env}
\paragraph{Domain-Grounded Service Modeling.}
Following \cite{agentskiller}, we use application domains as semantic anchors for subsequent service and environment generation.
Each domain specifies a coherent scope of user goals and activities, thereby preventing the generated components from becoming an arbitrary collection of unrelated concepts and operations.
Starting from a set of seed domains, we expand the domain pool into more specialized and role-specific scenarios and group related domains according to the activities they cover.
For example, the \textit{artist-booking} domain may cover activities such as identifying suitable artists, checking their availability, and arranging bookings for events.
Given the resulting domain group, we identify a primary user role and represent the information involved in the corresponding activities as entity schemas, each of which defines a type of object and its attributes.
We then construct an entity–relationship graph whose nodes represent entity types and whose edges encode the semantic relationships between them.
Based on coherent subsets of entities and relationships in this graph, we derive service blueprints that define the intended users, maintained information, and available functions of each service.
Each function is further described by its functional purpose, required and optional parameters, and returned information.
In the \textit{artist-booking} example, a \textit{booking coordinator} serves as the primary user, while \textit{artist}, \textit{client}, \textit{event}, and \textit{booking} constitute the main entity types.
The entity–relationship graph may associate each \textit{booking} with a \textit{client}, an \textit{artist}, and an \textit{event}, providing the semantic basis for functions that \textit{retrieve candidate artists}, \textit{check their availability}, and \textit{create bookings}.
After validation, each function is deterministically converted into a function-calling interface containing a function name and natural-language description, together with a JSON Schema object that specifies the names, types, descriptions, required fields, and enumeration constraints of its input parameters.
The resulting interfaces form the tool-call action space $\mathcal{A}_{\textrm{tool}}$ for each service.

\paragraph{From Service Blueprints to Stateful Environments.}
The service blueprints and function interfaces specify what each service should support, but they do not yet provide an environment in which the functions can be executed and their effects can persist.
We first instantiate each entity schema with concrete records and materialize the relationships between these records in a mock database.
The resulting records and relationships collectively define the initial external state $s_0^E$, after being checked against the corresponding schema and relational constraints.
In the \textit{artist-booking} example, the stateful database $s_0^E$ may contain concrete \textit{artists}, \textit{clients}, \textit{events}, and \textit{bookings}, with each \textit{booking} record referring to a specific \textit{client}, \textit{artist}, and \textit{event}.
A database alone, however, does not determine when a function may be invoked or how it changes the environment $s_t^E$.
We then define operational policies that specify each function’s preconditions, parameter semantics, execution constraints, state changes, and cross-function dependencies.
Using the function interfaces as callable contracts and the operational policies as behavioral specifications, we implement the function as executable server-side code in the MCP service.
Concretely, each implementation is a Python function registered under the name and input schema declared by the corresponding interface.
When invoked, it receives the supplied arguments, validates them against the schema, resolves the referenced records, enforces the applicable operational constraints, and performs the required read or write operations on the mock database.
It then converts the execution result or encountered error into the return format declared by the interface.
At runtime, the MCP server routes a tool action $a_t=(f_t,\mathbf{x}_t)\in\mathcal{A}_{\textrm{tool}}$ to the server-side implementation associated with $f_t$.
The function processes $\mathbf{x}_t$ according to the operational policy, updates $s_t^E$ when necessary, and returns an observation $o_{t+1}\in\mathcal{O}_{\textrm{tool}}$.

\paragraph{Grounded Task Synthesis and Execution Validation.}
Given a validated stateful environment, we next construct multi-step tasks whose required operations and information dependencies are grounded in the environment.
We first derive a tool-dependency graph from the function interfaces and operational policies.
Each node represents a function, while a directed edge indicates an execution dependency between two functions.
A valid path sampled on this graph naturally provides an executable structure that can be transformed into a multiple coordinated tool-call task.
To obtain diverse candidate structures, we sample paths from the graph by first drawing a starting node from a weighted probability distribution over eligible functions.
Because many user tasks follow a read-before-write pattern, first locating or inspecting target entities and then modifying them, we assign higher starting probabilities to query-type functions that retrieve information without changing $s_t^E$.
This preference is probabilistic rather than a hard constraint, allowing state-changing functions to remain possible starting points when their preconditions are already satisfied.

From the starting node, we extend the path on the graph, preserving dependency and parameter-flow constraints.
To improve complexity, we further construct tasks requiring longer or repeated tool use through two graph-grounded augmentation operations.
First, we join two paths when they share at least one function node.
The shared function provides an explicit semantic connection between the two sequences and serves as an anchor for organizing them under a unified task objective.
Second, we select an intermediate subpath and repeat it one or more times to model tasks that require iterating the same group of operations.
We retain the resulting sequence only if its function ordering, parameter flow, and overall task objective remain coherent.

Because graph connectivity alone does not guarantee that a function sequence corresponds to a real-world user request, we ground task synthesis in the domain semantics.
Specifically, we generate a natural-language task template and a unified user intent conditioned on the service blueprint, function interfaces, operational policies, and database summary.
The template must describe a single coherent user objective whose completion requires every function in the sampled sequence.
An independent evaluator then assesses each template in terms of motivation clarity, logical coherence, function coverage, naturalness, and specificity.
Templates that do not meet the quality threshold are discarded.

For each retained template, we determine the provenance of every required argument.
Arguments corresponding to facts or constraints known by the user are assigned to the user side of the task.
Identifiers and values produced during execution are linked to the preceding tool outputs.
We then instantiate the template with concrete records and parameter values consistent with the initial external state $s_0^E$.
Starting from $s_0^E$, we replay the grounded function sequence against the MCP service under $\mathcal{P}$, supplying user-known arguments from the instantiated task and propagating intermediate tool outputs to subsequent calls when required.
We retain only instances for which every action executes successfully and produces the intended observations and state transitions.
The validated execution is recorded as the reference tool trajectory $\tau_{\textrm{gold}}$, and its terminal external state is stored as $s_{\textrm{gold}}^E$.

\paragraph{Task-Conditioned User Simulation.}
To instantiate the user side of multi-turn interaction, we introduce an independent LLM-based user agent that collaborates with the evaluated agent throughout task execution.
The user agent is governed by a general interaction protocol and a private task handbook.
The interaction protocol requires the user agent to preserve the assigned intent, disclose information progressively, use only grounded values, and communicate from the user’s perspective without referring to functions or execution plans.
For each task, the private task handbook specifies the complete user intent $u$, the underlying motivation, the grounded information assigned to the user side of the task, and task-specific instructions governing when that information may be disclosed.
We additionally generate a concise initial request $o_0$ for the evaluated agent that expresses the user’s immediate need while leaving information intended for later clarification undisclosed.
During rollout, when the evaluated agent produces a natural-language response $y_t\in\mathcal{A}_{\textrm{resp}}$, the user agent generates the next observation $o_{t+1}\in\mathcal{O}_{\textrm{user}}$ conditioned on the interaction protocol, the private task handbook, and the visible dialogue.
When clarification is requested, the user agent provides the corresponding value from the task handbook.
Its visible dialogue contains only the natural-language messages exchanged between the two agents.
The generated response advances the interaction state $s_t^H$.
The interaction continues until the user agent indicates that its intent has been satisfied or identifies the request as outside the service scope.

\subsection{Evolving Tool Ecosystems Construction}
\label{app:evol-con}
\paragraph{Composable Capability Transformations.}
For each domain, we start from its executable base tool set $\mathcal{A}_{\textrm{tool}}^{(0)}=\mathcal{A}_{\textrm{tool}}$ and construct a domain-specific space of alternative capability realizations.
We represent each transformation as $\phi:P\mapsto R$, where both the source block $P$ and the target realization $R$ can contain one or more functions.
Both $P$ and $R$ are restricted to the same domain, and $R$ must preserve the capabilities represented by $P$ under a different interface structure.
The numbers of functions contained in $P$ and $R$ induce three general forms of transformation:
1) A one-to-one transformation refines access to an existing capability by renaming the function or parameters, adding parameters, or changing their formats;
2) A one-to-many transformation decomposes one function into several coordinated interfaces;
3) A many-to-one transformation integrates multiple functions into a unified interface.
For integration, the source block is further restricted to a contiguous subsequence observed in a validated task trajectory from the same domain.
This restriction grounds the merged interface in a coherent and executable workflow rather than an arbitrary combination of functions.
Given a source block $P$, we use an LLM conditioned on the domain’s service blueprint, function schemas, and operational policies to propose the target realization $R$ and its executable implementation.
Each candidate maintains explicit lineage links to the source functions it covers and is instantiated in the corresponding stateful environment for automated validation.
We retain a transformation only when its target realization $R$ preserves the required source capabilities $P$, task-relevant state changes, and externally consumed outputs under equivalent inputs.
Because transformations modify only local capability blocks, the evolution space can be expanded incrementally without reconstructing domains or services.

\paragraph{Capability-Preserving Ecosystem Sampling.}
The validated transformations define the alternative capability realizations available within each domain, but they do not determine which interfaces are exposed at each ecosystem version.
We organize these realizations through their source-to-target lineage and denote the active interface set at version $v$ by $\mathcal{A}_{\textrm{tool}}^{(v)}$.
The initial state $\mathcal{A}_{\textrm{tool}}^{(0)}$ contains all executable base interfaces, from which subsequent states are generated incrementally.
To construct $\mathcal{A}_{\textrm{tool}}^{(v)}$, we start from $\mathcal{A}_{\textrm{tool}}^{(v-1)}$ and apply a bounded number of validated transformations while leaving the remaining interfaces unchanged.
To emulate gradual interface update and backward compatibility, we sometimes introduce $R$ while retaining $P$, creating overlapping realizations that require agents to select among redundant interfaces.
In other cases, we deactivate the interfaces in $P$ after introducing $R$, requiring agents to access and compose the underlying capabilities through the new realization.
We accept a sampled state only when every capability in the base tool set remains accessible through its active interfaces, either directly or through a coordinated realization.
Using the lineage relations established above, we recursively verify this coverage and reject any transition that leaves a source capability without a valid realization.
For each accepted state, we adapt the domain’s operational policies and assemble the validated interface implementations into an executable MCP service.
To further simulate incomplete, compressed, or stale documentation in real-world tool ecosystem evolution, we optionally perturb selected function descriptions after constructing the executable service.
This produces controlled documentation shifts while keeping the available functions, parameter schemas, and execution semantics unchanged.

\paragraph{Task-Level Optimal Realization Search.}
Although the user intent and target state remain fixed, different ecosystem versions may provide different realizations for completing the same task.
For each task, we extract the ordered base-function sequence $(f_1,\ldots,f_n)$ from its validated trajectory $\tau_{\textrm{gold}}$.
At version $v$, a task realization is an ordered sequence of active realizations whose associated source blocks exactly cover this base-function sequence.
Under a specified efficiency criterion, we define the \textbf{reference path} as the feasible task realization with the minimum total criterion value within the constructed realization space.
Step counts tool calls, while cost sums per-call scores (1--10) assigned by an LLM based on each tool's definition and implementation to reflect its potential execution cost.
Let $\Phi^{(v)}(P)$ denote the realizations of source block $P$ available in $\mathcal{A}_{\textrm{tool}}^{(v)}$.
We search for the reference path by partitioning the base-function sequence into contiguous source blocks and selecting one active realization for each block:
\begin{align}
\rho_i^{(v,c)}
=
\underset{\substack{
P_1\oplus\cdots\oplus P_m=(f_1,\ldots,f_n)\\
R_j\in\Phi^{(v)}(P_j)
}}{\arg\min}
\sum_{j=1}^{m}c^{(v)}(R_j),
\end{align}
where $\oplus$ denotes sequence concatenation and $c^{(v)}(R_i)$ specifies the efficiency criterion for realization $R_i$.
We solve this sequence-covering problem through dynamic programming over the active realizations in the task’s domain.
In this paper, we use function-call count and execution cost as separate criteria to produce a step-minimal reference path and a cost-minimal reference path for each task and version.
These paths are stored as benchmark annotations for the evaluation protocol and are not prescribed as trajectories that the evaluated agent must reproduce.

\subsection{Evaluation Protocol and Metrics}
\label{app:metrics}
\paragraph{Outcome-based Task Success.}
We first evaluate task completion from environment outcomes.
For each task $i$ and version $v$, the corresponding MCP service is initialized with the task’s external state $s_{0,i}^E$, and the agent trajectory $\tau_i^{(v)}$ produces a terminal state $s_{T,i}^{E,(v)}$.
Let $\Delta(s,s')$ denote the set of database changes between two external states, including inserted, updated, and deleted records.
We obtain the required and agent-produced changes as:
\begin{align}
    \Delta_i^{\mathrm{gold}}
=
\Delta(s_{0,i}^E,s_{\mathrm{gold},i}^E),
\qquad
\Delta_i^{(v)}
=
\Delta(s_{0,i}^E,s_{T,i}^{E,(v)}).
\end{align}
A task is considered successful when the agent-produced changes cover all changes required by the validated gold execution:
\begin{align}
    S_i^{(v)}
=
\mathcal{R}(\tau_i^{(v)},u_i)
=
\mathbb{I}\!\left[
\Delta_i^{\mathrm{gold}}
\subseteq
\Delta_i^{(v)}
\right].
\end{align}
For the \(N_v\) tasks evaluated at version \(v\), we report the success rate:
\begin{align}
    \mathrm{SR}^{(v)}
=
\frac{1}{N_v}
\sum_{i=1}^{N_v}S_i^{(v)}.
\end{align}

\paragraph{Tool Retrieval Quality.}
Beyond task completion, we evaluate whether the agent can accurately identify tools required by the current task from the complete action space.
For task $i$ at version $v$, let $\widehat{\mathcal A}_i^{(v)}\subseteq\mathcal A_{\textrm{tool}}^{(v)}$ denote the final tool set exposed to the agent after retrieval.
We use $\mathcal A_i^{(v)}\subseteq\mathcal A_{\textrm{tool}}^{(v)}$ to denote the tools contained in the reference path.
The task-level retrieval P, R, and F1 are defined as:
\begin{align}
\mathrm{P}_i^{(v)}
=
\frac{
\left|\widehat{\mathcal A}_i^{(v)}\cap\mathcal A_i^{(v)}\right|
}{
\left|\widehat{\mathcal A}_i^{(v)}\right|
},
\qquad
\mathrm{R}_i^{(v)}
=
\frac{
\left|\widehat{\mathcal A}_i^{(v)}\cap\mathcal A_i^{(v)}\right|
}{
\left|\mathcal A_i^{(v)}\right|
},
\qquad
\mathrm{F1}_i^{(v)}
=
\frac{
2\mathrm{P}_i^{(v)}\mathrm{R}_i^{(v)}
}{
\mathrm{P}_i^{(v)}+\mathrm{R}_i^{(v)}
}.
\end{align}

\paragraph{Success-weighted Path Efficiency.}
We define Success-weighted Path Efficiency (SPE) to measure how efficiently an agent completes tasks relative to the reference paths.
For task \(i\) at version \(v\), let \(\mathcal{W}_i^{(v)}\) denote the reference-path weight and \(\widehat{\mathcal{W}}_i^{(v)}\) the weight of the agent’s tool-call trajectory under the same criterion.
SPE is defined as
\begin{align}
\mathrm{SPE}^{(v)}
=
\frac{1}{N_v}
\sum_{i=1}^{N_v}
S_i^{(v)}
\min\left(
\frac{\mathcal{W}_i^{(v)}}
{\widehat{\mathcal{W}}_i^{(v)}},
1
\right).
\end{align}
Using function-call count and execution cost as the path weight yields Step-SPE and Cost-SPE.

\paragraph{Improvement Discovery and Degradation Adaptation.}
To evaluate whether the agent can adapt to the evolving frontier, we compare the reference-path weight of each task across adjacent ecosystem versions.
We define the tasks whose optimal execution decreases or increases as:
\begin{align}
\mathcal{D}_{\downarrow}^{(v)}
=
\left\{
i \mid
\mathcal{W}_{i}^{(v)}
<
\mathcal{W}_{i}^{(v-1)}
\right\},
\qquad
\mathcal{D}_{\uparrow}^{(v)}
=
\left\{
i \mid
\mathcal{W}_{i}^{(v)}
>
\mathcal{W}_{i}^{(v-1)}
\right\}.
\end{align}
The Improvement Discovery Rate (IDR) and Degradation Adaptation Rate (DAR) are defined as:
\begin{align}
\small
\mathrm{IDR}^{(v)}
=
\frac{1}{\left|\mathcal{D}_{\downarrow}^{(v)}\right|}
\sum_{i\in\mathcal{D}_{\downarrow}^{(v)}}
S_i^{(v)}
\min\left(
\frac{\mathcal{W}_{i}^{(v)}}
{\widehat{\mathcal{W}}_{i}^{(v)}},
1
\right),
\mathrm{DAR}^{(v)}
=
\frac{1}{\left|\mathcal{D}_{\uparrow}^{(v)}\right|}
\sum_{i\in\mathcal{D}_{\uparrow}^{(v)}}
S_i^{(v)}
\min\left(
\frac{\mathcal{W}_{i}^{(v)}}
{\widehat{\mathcal{W}}_{i}^{(v)}},
1
\right).
\end{align}
IDR measures whether the agent discovers newly shortened or cheaper execution paths, whereas DAR measures whether it remains effective when the optimal path becomes longer or more costly.
Using function-call count or execution cost as the path weight yields Step-IDR/DAR and Cost-IDR/DAR.

\begin{table}[t]
\centering
\begin{minipage}[t]{0.45\textwidth}
    \vspace{0pt}
    \centering
    \renewcommand\arraystretch{1.}
    \setlength{\tabcolsep}{4pt}
    \caption{\small
    \textbf{Statistics of evolving tool ecosystems.}
    Tool scale across six ecosystem states and the corresponding interface changes.
    }
    \label{tab:ecosystem_statistics}
    \vspace{-0.08in}
    \resizebox{\linewidth}{!}{
    \begin{tabular}{c|cc|ccc}
    \toprule
    \multirow{2}{*}{\textbf{Ecosystem}}
    & \multicolumn{2}{c|}{\textbf{Tool Scale}}
    & \multicolumn{3}{c}{\textbf{Interface Changes}} \\
    \cmidrule{2-3} \cmidrule{4-6}
    & \textbf{\# Tools}
    & \textbf{Avg.}
    & \textbf{Added}
    & \textbf{Retired}
    & \textbf{Modified} \\
    \midrule
    $v_1$ & 892 & 21.24 & -- & -- & -- \\
    $v_2$ & 841 & 20.02 & 110 & 161 & 176 \\
    $v_3$ & 837 & 19.93 & 196 & 200 & 147 \\
    $v_4$ & 799 & 19.02 & 125 & 163 & 150 \\
    $v_5$ & 858 & 20.43 & 179 & 120 & 155 \\
    $v_6$ & 890 & 21.19 & 166 & 134 & 165 \\
    \bottomrule
    \end{tabular}
    }
\end{minipage}
\hfill
\begin{minipage}[t]{0.53\textwidth}
    \vspace{0pt}
    \centering
    \includegraphics[width=\linewidth]
        {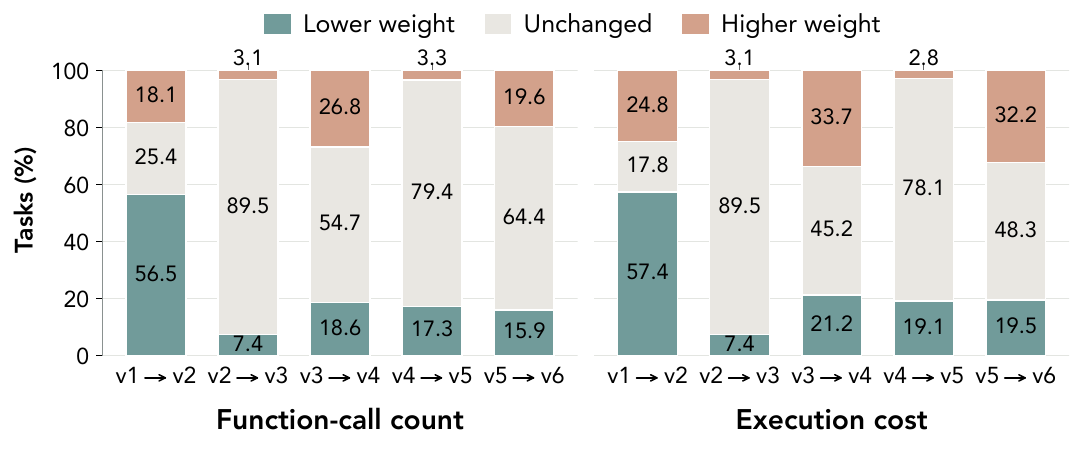}
    \vspace{-0.23in}
    \captionof{figure}{\small
    \textbf{Reference-path shifts across transitions.}
    Task distributions by function-call count and execution cost.
    }
    \label{fig:efficiency_frontier}
\end{minipage}
\end{table}

\subsection{Benchmark Statistics and Characteristics}
\label{app:statistics}
Overall, {\bench} contains 1,588 tasks spanning 42 domains, with 1,191 tasks in the training set and 397 in the test set; across six ecosystem states, this yields 9,528 task–ecosystem instances.
As shown in Tab.~\ref{tab:ecosystem_statistics}, the aggregate tool scale remains comparable across states, whereas Fig.~\ref{fig:efficiency_frontier} reveals non-monotonic changes in reference-path weights across transitions.
The different distributions under function-call count and execution cost further indicate that interaction length and execution expense capture complementary dimensions of efficiency.
Tab.~\ref{tab:domain_statistics} further details the task counts and mean tool-call lengths of the original gold (reference) trajectories for each domain and split.

\begin{table}[t]
\centering
\small
\setlength{\tabcolsep}{3pt}
\renewcommand{\arraystretch}{1.08}
\vspace{-1cm}
\caption{\small \textbf{Domain-level task statistics.}
Mean calls are computed from the original gold trajectories, before ecosystem-specific interface transformations.}
\label{tab:domain_statistics}
\begin{tabular*}{\linewidth}{@{\extracolsep{\fill}}lrrrrr@{}}
\toprule
\multirow{2}{*}{\textbf{Domain}}
& \multicolumn{3}{c}{\textbf{Tasks}}
& \multicolumn{2}{c}{\textbf{Mean gold calls}} \\
\cmidrule(lr){2-4}\cmidrule(l){5-6}
& \textbf{Train} & \textbf{Test} & \textbf{Total}
& \textbf{Train} & \textbf{Test} \\
\midrule
AgriculturalEngineeringTrainingServer & 34 & 11 & 45 & 4.32 & 4.36 \\
AgriculturalSkillsTrainingHub & 32 & 10 & 42 & 4.69 & 5.00 \\
ArtistBookingAgency & 13 & 4 & 17 & 4.54 & 4.75 \\
ArtistCareerTracker & 62 & 20 & 82 & 5.45 & 5.55 \\
ArtistContractComplianceAdvisor & 53 & 18 & 71 & 5.60 & 5.50 \\
ArtistTeachingStudio & 42 & 14 & 56 & 4.67 & 4.79 \\
ClassroomSafetyAnalyzer & 26 & 9 & 35 & 4.92 & 4.89 \\
ClinicalEfficacyAnalyzer & 70 & 24 & 94 & 4.93 & 5.04 \\
ClinicalProcedureManager & 10 & 4 & 14 & 4.30 & 4.50 \\
CommunityFacilitySchedulingCoordinator & 5 & 2 & 7 & 4.80 & 5.00 \\
ConstructionProjectRiskServer & 43 & 14 & 57 & 5.16 & 5.50 \\
CourseComplianceManager & 11 & 4 & 15 & 5.18 & 4.50 \\
CurriculumResourceIntegrator & 58 & 19 & 77 & 5.14 & 4.89 \\
EngineeringSpecificationComplianceServer & 50 & 16 & 66 & 5.74 & 5.56 \\
FarmEquipmentServiceNetwork & 37 & 12 & 49 & 5.43 & 5.58 \\
FirstResponderCommunityResourceCoordination & 17 & 6 & 23 & 4.88 & 5.33 \\
FirstResponderFieldOpsCommunicationAndMonitoring & 14 & 5 & 19 & 4.43 & 4.40 \\
FirstResponderIncidentReportingAndSceneMgmt & 7 & 2 & 9 & 6.43 & 6.50 \\
FirstResponderTrainingAndEquipmentCertification & 2 & 0 & 2 & 4.00 & -- \\
FleetServiceCoordinator & 33 & 11 & 44 & 5.18 & 5.27 \\
GrantAndResourceComplianceTracker & 1 & 0 & 1 & 4.00 & -- \\
InfrastructureComplianceManager & 62 & 20 & 82 & 5.03 & 5.25 \\
InfrastructureMonitoringHub & 46 & 16 & 62 & 5.35 & 5.25 \\
LivestockHealthMonitor & 20 & 7 & 27 & 4.80 & 5.00 \\
MedicalDeviceEngineeringServer & 84 & 28 & 112 & 5.65 & 5.79 \\
MobileServiceDeploymentOptimizer & 13 & 4 & 17 & 4.38 & 4.00 \\
MobilityPlatformIntegrator & 34 & 13 & 47 & 5.12 & 5.08 \\
PharmaceuticalTreatmentOptimizer & 12 & 4 & 16 & 5.42 & 5.25 \\
PolicyImpactReviewer & 28 & 10 & 38 & 4.71 & 4.70 \\
PolicyImplementationAndRiskAdvisor & 16 & 5 & 21 & 4.56 & 4.40 \\
PrecisionCropManagementAdvisor & 56 & 19 & 75 & 5.14 & 5.05 \\
PublicAssistanceWorkflowManager & 3 & 1 & 4 & 4.33 & 4.00 \\
PublicEmployeeSupportPortal & 14 & 5 & 19 & 5.07 & 5.00 \\
PublicInfrastructurePlanner & 78 & 26 & 104 & 4.87 & 4.85 \\
RiskMonitorCoordinator & 49 & 16 & 65 & 5.16 & 4.94 \\
ServiceProviderCompensationAndSupportManager & 18 & 6 & 24 & 5.00 & 5.33 \\
ServiceProviderManager & 4 & 1 & 5 & 6.00 & 5.00 \\
ServiceQualityAndInsightGenerator & 19 & 6 & 25 & 4.16 & 4.00 \\
SmartInfrastructureMonitoringServer & 3 & 1 & 4 & 4.00 & 4.00 \\
StudentSupportProgramConnector & 4 & 1 & 5 & 5.75 & 5.00 \\
SupplyChainSpecificationVerifier & 4 & 2 & 6 & 5.75 & 5.00 \\
TeacherManager & 4 & 1 & 5 & 4.75 & 4.00 \\
\midrule
\textbf{Overall} & \textbf{1,191} & \textbf{397} & \textbf{1,588} & \textbf{5.10} & \textbf{5.11} \\
\bottomrule
\end{tabular*}
\end{table}

\section{Details of Skill-based Baselines}
\label{app:baselines}

\begin{itemize}
    \item \textbf{AutoSkill} \citep{autoskill} extracts reusable skills from user inputs to support continual improvement without updating model parameters.
    It maintains a skill library by adding distinct capabilities and merging refinements into existing skills, while filtering unsuitable or redundant candidates.
    For subsequent requests, relevant skills are retrieved and injected into the agent's context to guide its responses.

    \item \textbf{Trace2Skill} \citep{trace2skill} distills experience from successful and failed execution trajectories into a reusable skill directory.
    Separate analysts process trajectories in parallel to propose skill patches, which are hierarchically consolidated to resolve conflicts and retain generalizable procedures.
    The resulting skill guides subsequent execution without parameter updates or retrieval of individual trajectory memories.

    \item \textbf{EvoSkill} \citep{evoskill} discovers and refines reusable skills through iterative failure analysis.
    A proposer diagnoses unsuccessful executions and suggests new or revised skills, which a skill-builder turns into structured skill folders containing procedural instructions and optional supporting resources.
    Candidate agent programs are evaluated on held-out validation data, and the highest-scoring programs are retained for further evolution while the underlying model remains frozen.

    \item \textbf{SkillOpt} \citep{skillopt} optimizes a natural-language skill document while keeping the target model fixed.
    A separate optimizer model converts scored execution trajectories into textual edits under a limited edit budget, and each update is accepted only if it improves held-out validation performance.
    Rejected edits provide feedback for future proposals, while consolidation across epochs preserves longer-term lessons.
\end{itemize}

For all baselines, we directly adapt their official implementations to {\bench} without revising any modules and use DeepSeek-V3.2 as the common backbone.

\section{Details of {\ours}}
\label{app:method-detial}
\paragraph{Skill Structure and Initialization.}
{\ours} organizes procedural knowledge into a three-level skill library.
Planning skills describe task-level workflows, while atomic skills encode reusable operations with explicit tool dependencies.
Tool skills provide invocation guidance for individual tools.
Each atomic skill stores a name, a description, reusable usage guidance, and its required tools.
The skill graph links planning skills to associated atomic skills and atomic skills to the tool skills corresponding to their dependencies.
At $v=1$, we initialize the library following \citep{skillx}.
Planning and atomic skills are extracted from training trajectories that pass environment-state evaluation.
Planning extraction groups tool calls by their underlying subgoals, removing exploration, debugging, and failed attempts while preserving execution order.
Atomic extraction focuses on one specified tool at a time, replacing task-specific values with reusable parameters and retaining relevant coordination with other tools.
Tool-skill summarization additionally combines tool schemas with repeated tool-call failure evidence to identify invocation constraints and recovery procedures.
The corresponding prompts used in this paragraph are detailed in Appx.~\ref{app:skill_initialization_prompts}.

\paragraph{Incremental Knowledge Maintenance.}
After initialization, planning skills remain fixed, while {\ours} incrementally maintains atomic and tool skills using newly verified training trajectories.
Given task metadata, the evaluation outcome, a trajectory summary, and retrieved skills, the manager proposes \texttt{add}, \texttt{modify}, \texttt{keep}, or \texttt{skip} decisions.
Updates prioritize reusable workflow guidance, including tool dependencies and recovery procedures, while avoiding duplicate skills or changes unsupported by new experience.
Cost-aware guidance preserves all task-required actions and favors cheaper alternatives only when trajectory evidence establishes the same preconditions and state effects.
It may also identify redundant calls or known-invalid retries.
The manager inherits both active and inactive records from the preceding ecosystem state and merges updates by skill name, with newer records replacing older ones.
A tool skill is active when its associated tool exists in the current catalog, whereas an atomic skill is active only when all its declared tool dependencies are available.
Missing dependencies are recorded, and inactive skills are retained for reactivation when their required tools become available again.
Retrieval indices and the skill graph are rebuilt over the active subset.
An auxiliary distillation step learns retrieval-policy skills from successful episodes in which newly retrieved tools were subsequently invoked.
These skills provide conditional guidance for retrieval timing, query formulation, or result assessment.
They remain outside the hierarchical skill graph and do not directly change tool visibility.
Prompts are detailed in Appx.~\ref{app:knowledge_maintenance_prompts}.

\paragraph{Initial Capability Provisioning.}
At the start of each episode, {\ours} retrieves the top five planning skills using the initial user request and asks an LLM to rewrite them into a task-specific provisional plan $\hat p$.
This produces a structured list of actionable steps without modifying the stored planning skills.
For each step, the manager retrieves up to five atomic skills subject to a similarity threshold.
Candidates linked to the retrieved planning skills receive priority, followed by deduplication and maximal marginal relevance (MMR) to balance relevance and diversity.
An LLM then selects at most 8 atomic skills based on the request and provisional plan.
Their linked tool skills are attached to provide detailed invocation guidance, together forming the initial procedural context $k_0^{(v)}$.
To complement skill-based retrieval, the manager directly retrieves the top three tools for each plan step.
The union of these tools and the selected skills' declared dependencies is intersected with the current tool catalog to obtain the initial visible set $\widehat{\mathcal A}_0^{(v)}$.
The visible tools and procedural context constitute the initial task-specific view $b_0^{(v)}$.
The skill context is supplied as a system message at each turn without appending repeated copies to the dialogue history.
It instructs the task agent to apply skills only when relevant and to rely on the actual session tool list when deciding which tools can be called.
Prompts for this process are shown in Appx.~\ref{app:initial_provisioning_prompts}.

\paragraph{On-Demand Capability Provisioning.}
During execution, the task agent can call \texttt{retrieve} \texttt{\_tools} with a focused capability query to retrieve up to five additional tools when its current tool set is insufficient.
For gaps requiring further investigation, it can call \texttt{explore\_capabilities} to delegate a focused search to the manager, with at most three delegations per episode.
Each request specifies the task goal and missing capability, optionally including previously failed attempts.
The manager also receives recent user messages and tool errors, the provisional plan, and the names of already visible tools and injected skills.
It searches the active skill library and full tool catalog, using skill bodies and graph links to recover relevant procedural guidance.
During this investigation, the manager cannot execute service tools or communicate with the user, and it must inspect each candidate tool's schema before recommending it.
If a referenced tool no longer exists, it searches for a semantically equivalent replacement and explicitly reports the substitution.
Among equivalent options, the prompt favors lower cost or latency.
The investigation operates within an action budget and ends as soon as the delegated request can be answered.
The manager returns a compact report specifying recommended tools and an actionable workflow, together with relevant pitfalls and rejected alternatives.
At most 8 tools are returned per investigation, and a programmatic check admits only tools present in the current catalog into the task agent's visible set.
When summarization would lose parameter-level or ordering details, the manager can additionally provide the full text of at most two skills.
Skill-body delivery is limited to 4,000 characters cumulatively per episode, and each new added-skills block replaces the previous block.
Providing a skill does not automatically expose its referenced tools, which must be inspected and admitted separately.
These outputs jointly update the available tools $\widehat{\mathcal A}_t^{(v)}$ and procedural context $k_t^{(v)}$ as execution reveals new capability requirements.
Please refer to Appx.~\ref{app:on_demand_provisioning_prompts} for prompts used in this paragraph.

\section{Prompts}
\label{app:prompts}

\subsection{Skill Structure and Initialization}
\label{app:skill_initialization_prompts}

\subsubsection{Planning Skill Extraction}
\label{app:planning_extraction_prompt}

\begin{tcolorbox}[
    breakable,
    colback=black!5!white,
    colframe=black!60!black,
    left=1ex, right=1ex,
    top=0.5ex, bottom=0.5ex,
    fontupper=\small\ttfamily,
    fontlower=\small\ttfamily,
    title=\textbf{Planning Skill Extraction}
]
\textbf{System Prompt:}

You are a \textbf{Planning Expert}.
Your job is to analyze an agent's API interaction history and the user's task, then distill them into a concise, reusable plan.
This plan should serve as a reference for handling similar tasks more effectively in the future.

\medskip
\textbf{Objectives:}
\begin{enumerate}
    \item \textbf{Understand Capabilities.}
    Analyze the recorded API calls to identify the actual functional capabilities demonstrated.
    \item \textbf{Abstract into a Plan.}
    For each feasible task supported by those capabilities, produce a concise, reusable step-by-step plan that can be applied to similar tasks.
\end{enumerate}

\medskip
\textbf{Planning Creation Rules:}
\begin{enumerate}
    \item \textbf{Focus.}
    Do not simply restate each API function step-by-step using technical jargon.
    Instead, describe the underlying sub-goal behind each action segment.

    \item \textbf{Remove Non-Essential Steps.}
    Exclude capability exploration, debugging, and failed steps.

    \item \textbf{Reusability.}
    The plan must be precise enough for other models to reuse.

    \item \textbf{Conciseness.}
    Merge steps from the interaction history that share the same objective into a single sub-step in the plan.
    Use a compact writing style for each sub-step, while listing the key APIs involved in that step (one or more).
    Do not omit any critical, potentially required API keys.
\end{enumerate}

\medskip
\textbf{Output Format:}

For each task, output exactly one plan and follow this format strictly:

\smallskip
\noindent
<plan>\\
\# step 1: A natural, specific, concise sub-task goal; key APIs used (one or more).\\
\# step 2: ...\\
...\\
</plan>

\medskip
\textbf{Good Examples:}

\{PLAN\_EXAMPLE\}

\medskip
\textbf{Checklist Before Finalizing:}
\begin{itemize}
    \item \textbf{Reusability.}
    Ensure no critical steps are missing and the step order is correct.
    \item \textbf{Conciseness.}
    Confirm there are no redundant or unnecessary steps.
    \item \textbf{Agent-centered.}
    Make sure the plan reads like actionable instructions that other models can reliably follow.
\end{itemize}

\medskip
\textbf{User Prompt:}

\textbf{User Task:}

\{USER\_TASK\}

\medskip
\textbf{Trajectory:}

\{TRAJECTORY\}

\end{tcolorbox}

\subsubsection{Atomic Skill Extraction}
\label{app:atomic_extraction_prompt}

\begin{tcolorbox}[
    breakable,
    colback=black!5!white,
    colframe=black!60!black,
    left=1ex, right=1ex,
    top=0.5ex, bottom=0.5ex,
    fontupper=\small\ttfamily,
    fontlower=\small\ttfamily,
    title=\textbf{Atomic Skill Extraction}
]
\textbf{System Prompt:}

An agent system is provided with a \textbf{Skill Library} and has tried to solve the task multiple times with a successful solution.
Review the task-solving attempt and extract generalizable skills.

\medskip
\textbf{Inputs Description:}
\begin{itemize}
    \item \textbf{User Task}
    \item \textbf{Trajectory:}
    A record of an agent's interactions successfully with the environment as it attempts to complete a user task.
    \item \textbf{Skill Library:}
    A collection of all currently available skills that can be directly reused.
    \item \textbf{Specific-Tool:}
    Given a specific tool, extract only one reusable skill for the specified tool.
\end{itemize}

\medskip
\textbf{Skill Definition Rule:}

Skill is a dictionary with four keys:
\texttt{name}, \texttt{document}, \texttt{content}, and \texttt{tools}.
\begin{enumerate}
    \item \texttt{name}: the specific tool's name.
    \item \texttt{document}: the tool's functionality, the key parameters, the final output of the skill and any important notes.
    \item \texttt{content}: the tool's usage examples, and examples of combining it with other tools (if applicable).
    \item \texttt{tools}: the key tools used in the \texttt{content} (list).
\end{enumerate}

\begin{itemize}
    \item The skill is centered around a specific tool, describing its core functionality, important notes, and common usage examples.
    \item Explicitly declare the key parameters and the final output data types using type hints.
    Example: \texttt{Parameters: param: str; Outputs: output: dict:}
    \item Include detailed description of the skill with input and output explanation.
    \item The skill should not be similar to the existing skills in the skill library.
    \item The parameters used in \texttt{content} must be reusable instead of hard-coded values (e.g., specific email address ``jay@gmail.com'').
    \item The usage examples of \texttt{content} may involve one or more tool uses.
    \item The \texttt{document} must clearly and thoroughly document all relevant details of the specific tool use.
    \item Never call other skills from the skill library or any previously defined skills.
    \item Do not import any Python packages.
    \item Avoid a functional style and Python code style; there's no need to use return.
\end{itemize}

\medskip
\textbf{Update Existing Skills:}

Your goal is to ensure the system retains actionable skill that helps it behave correctly in the future.

You have three options: \textbf{[modify, add, keep]}
\begin{itemize}
    \item \textbf{modify:}
    Revise an existing skill to make it more effective (e.g., improving documents).
    Only change \texttt{content} when necessary, and ensure the resulting skill remains broadly general-purpose.
    \item \textbf{add:}
    Introduce a new skill only when existing skill library missing the the specified tool.
    \item \textbf{keep:}
    Preserve the skill unchanged when there are no clear issues.
    \item If the \textbf{Skill Library} is empty, you may only use \textbf{add}.
    \item \textbf{modify} and \textbf{keep} are only valid for skills that already exist in the provided \textbf{Skill Library}.
    \item For \textbf{modify}, \texttt{modified\_from} must exactly match an existing skill name from the provided \textbf{Skill Library}.
    \item For \textbf{keep}, \texttt{skill\_name} must exactly match an existing skill name from the provided \textbf{Skill Library}.
\end{itemize}

\textbf{Common actions:}
\begin{itemize}
    \item add a new skill
    \item update a skill's usage instructions/documentation
    \item revise a skill's variable/parameter definitions to make it more generalizable
    \item keep a skill unchanged
    \item ...
\end{itemize}

\medskip
\textbf{Requirements for Each Skill That Is Modified or Added:}
\begin{itemize}
    \item \textbf{Avoid duplication:}
    If a skill library is provided, do not add new skills that are similar to existing ones---use \textbf{keep} or \textbf{modify} instead.
    \item \textbf{Ensure domain specificity:}
    The skill must contain domain-specific tool.
    \item \textbf{Specific-Tool guided extraction:}
    Only focus on the specified tool in the trajectory when extracting skills.
\end{itemize}

\medskip
\textbf{Good Skill Example:}

\{GOOD\_SKILL\_EXAMPLE\}

\medskip
\textbf{Output Format:}

you will finish by returning in this JSON format as follows:
\begin{verbatim}
{OUTPUT_FORMAT_EXAMPLE}
\end{verbatim}

Note that your updated skills may not need to cover all the options.
You can only use one type of updates or choose to remain all skills unchanged.

Return a JSON array only.
Each array item must be an object.
\texttt{option} must be one of \texttt{modify}, \texttt{add}, or \texttt{keep}.
For \texttt{modify} and \texttt{add}, \texttt{skill} must be an object with keys \texttt{name}, \texttt{document}, \texttt{content}, and \texttt{tools}, and \texttt{tools} must be a list of strings.
If the provided \textbf{Skill Library} is empty, do not output any \texttt{modify} or \texttt{keep} item.

\medskip
\textbf{Checklist Before Finalizing:}
\begin{itemize}
    \item \textbf{Reusability.}
    Ensure no critical steps are missing, each skill is modular, all parameters are abstract rather than specific.
    \item \textbf{Optimality.}
    Ensure each skill meets the required definition standards.
    \item \textbf{Agent-centered.}
    Add helpful notes in each skill to guide other models in using it correctly.
    \item \textbf{Specific-Tool focus.}
    Whether the extracted skill doesn't center around this Tool?
\end{itemize}

\medskip
\textbf{User Prompt:}

\textbf{User Task:}

\{USER\_TASK\}

\medskip
\textbf{Trajectory:}

\{TRAJECTORY\}

\medskip
\textbf{Skill Library:}

\{SKILL\_LIBRARY\}

\medskip
\textbf{Specific-Tool:}

\{TARGET\_TOOL\_NAME\}

\end{tcolorbox}

\subsubsection{Tool Skill Extraction}
\label{app:tool_skill_extraction_prompt}

\begin{tcolorbox}[
    breakable,
    colback=black!5!white,
    colframe=black!60!black,
    left=1ex, right=1ex,
    top=0.5ex, bottom=0.5ex,
    fontupper=\small\ttfamily,
    fontlower=\small\ttfamily,
    title=\textbf{Tool Skill Extraction}
]
\textbf{System Prompt:}

You are constructing a reusable tool skill for an agent skill library.

Your task is to read the tool schema and repeated rollout evidence about a single tool-use failure pattern, then write ONE reusable tool skill that helps future agents avoid or recover from that failure.

\medskip
\textbf{Goal:}

Produce a tool-centric skill that captures:
\begin{enumerate}
    \item what the tool is for,
    \item the important input/output details,
    \item the specific failure pattern seen in the evidence,
    \item the preconditions or parameter constraints needed to avoid that failure,
    \item a generic recovery pattern.
\end{enumerate}

\medskip
\textbf{Constraints:}
\begin{itemize}
    \item Output exactly one JSON object inside \texttt{<skill>...</skill>}.
    \item The JSON object must contain exactly these keys:
    \texttt{name}, \texttt{document}, \texttt{content}.
    \item \texttt{name} must be the target tool name, not a new abstract name.
    \item \texttt{document} should describe the tool, key parameters, outputs, and important notes from both schema and evidence.
    \item \texttt{content} should be reusable natural-language usage examples, not Python code.
    \item If the rollout evidence shows the tool is actually used in a way that differs from the schema, explicitly clarify that usage pattern in the skill instead of assuming the schema is the full story.
    \item Keep the skill generic and reusable; do not narrate a specific user story.
\end{itemize}

\medskip
\textbf{Output Format:}
\begin{verbatim}
<skill>
{
  "name": "tool_name",
  "document": "...",
  "content": "..."
}
</skill>
\end{verbatim}

\medskip
\textbf{User Prompt:}

\textbf{Tool Schema:}

\{TOOL\_SCHEMA\}

\medskip
\textbf{Repeated Failure Evidence:}

\{FAILURE\_EVIDENCE\}

\end{tcolorbox}

\subsection{Incremental Knowledge Maintenance}
\label{app:knowledge_maintenance_prompts}

\subsubsection{Incremental Skill Maintenance}
\label{app:skill_maintenance_prompt}

\begin{tcolorbox}[
    breakable,
    colback=black!5!white,
    colframe=black!60!black,
    left=1ex, right=1ex,
    top=0.5ex, bottom=0.5ex,
    fontupper=\small\ttfamily,
    fontlower=\small\ttfamily,
    title=\textbf{Incremental Skill Maintenance}
]
\textbf{System Prompt:}

You are updating a graph-backed skill library from one rollout session.

You will be given:
\begin{itemize}
    \item task metadata and reward outcome
    \item a summarized trajectory
    \item planning skills retrieved into context
    \item atomic skills retrieved into context
    \item tool skills retrieved/attached into context
\end{itemize}

Your job is to decide which tool skills and atomic skills should be kept, modified, or newly added.
The resulting skills are for a future agent to reuse while solving user tasks.
They should describe agent-side workflow guidance, tool usage constraints, recovery behavior, and reusable execution patterns.
Do not turn user behavior or conversation style into a skill unless it directly constrains how the agent should act.

\medskip
\textbf{Prioritization:}
\begin{itemize}
    \item Prefer updates that capture non-trivial, reusable agent guidance.
    \item Prefer sessions that reveal complex workflow, multi-step coordination, hidden dependency, recovery behavior, or easy-to-miss constraints.
    \item Prefer guidance that would help a future agent avoid mistakes, avoid dead ends, or handle a subtask that is somewhat hard rather than obvious.
    \item A successful session can still justify an update if it demonstrates a non-obvious workflow or a subtle but important control rule.
    \item If a subtask was solved in a simple, direct, obvious way and the session adds no meaningful new guidance, prefer ``keep'' or ``skip''.
\end{itemize}

\medskip
\textbf{High-Value Signals:}
\begin{itemize}
    \item lookup -> act, validate -> execute, preview -> confirm, or retrieve ID -> perform update patterns
    \item tool errors, retries, repair steps, argument correction, format correction, or fallback handling
    \item multi-tool coordination, order dependence, or context-building before action
    \item cases where the correct action is not just ``call the tool directly''
    \item reusable guidance for difficult or failure-prone subtasks
\end{itemize}

\medskip
\textbf{Low-Value Signals:}
\begin{itemize}
    \item merely restating what a tool does
    \item retelling the task or conversation
    \item documenting a one-step happy-path call that is already obvious from the tool itself
    \item adding generic advice with no concrete workflow value
    \item overfitting to task-specific facts instead of extracting reusable guidance
\end{itemize}

\medskip
\textbf{Rules:}
\begin{itemize}
    \item Tool skills are one-tool-one-skill.
    \item If a tool already has a tool skill in context, you may only use option ``modify'' or ``keep'' for that tool.
    \item If a tool does not yet have a tool skill in context, you may only use option ``add'' or ``skip''.
    \item For tool skills, do NOT output a \texttt{tools} field; the system will set it automatically.
    \item Tool skill updates should focus on agent-facing pitfalls, prerequisites, verification steps, confirmation semantics, argument traps, and common recovery patterns for that tool.
    \item Tool skills should NOT mainly be API summaries, generic capability descriptions, or obvious parameter recitations.
    \item For existing atomic skills in context:
    \begin{itemize}
        \item use ``modify'' only if the skill needs a real update
        \item use ``keep'' if no update is needed
    \end{itemize}
    \item You may use atomic ``add'' only for a genuinely new reusable capability that is not just a clarification of an existing atomic skill.
    \item Atomic skill updates should focus on reusable multi-step subtask workflow: sequencing, branching, validation, retry, recovery, and tool coordination.
    \item Atomic skills should NOT mainly capture trivial single-call behavior or task retelling.
    \item Prioritize reusable agent workflow guidance over task-specific narration.
    \item Do NOT force updates.
    If the session does not contribute meaningful new reusable guidance, choose keep/skip.
    \item Do NOT propose any planning skill updates here.
\end{itemize}

\medskip
\textbf{Output Format:}

Output exactly one JSON object inside \texttt{<proposal>...</proposal>} with these keys:
\begin{itemize}
    \item \texttt{confidence}: number between 0 and 1
    \item \texttt{reasoning\_summary}: string
    \item \texttt{tool\_skill\_updates}: list[object]
    \item \texttt{atomic\_skill\_updates}: list[object]
\end{itemize}

Each tool update object must have:
\begin{itemize}
    \item \texttt{option}: ``add'' | ``modify'' | ``keep'' | ``skip''
    \item \texttt{tool\_name}: string
    \item \texttt{rationale}: string
    \item \texttt{skill}: object or null
\end{itemize}

Each atomic update object must have:
\begin{itemize}
    \item \texttt{option}: ``add'' | ``modify'' | ``keep'' | ``skip''
    \item \texttt{skill\_name}: string
    \item \texttt{modified\_from}: string or ""
    \item \texttt{rationale}: string
    \item \texttt{skill}: object or null
\end{itemize}

For add/modify skill objects:
\begin{itemize}
    \item atomic skill must contain \texttt{name/document/content/tools}
    \item tool skill must contain \texttt{name/document/content}
\end{itemize}

\medskip
\textbf{Cost-Aware Skill Guidance:}

Correctness is the primary objective.
Preserve every necessary state change, required precondition, confirmation, and task-required verification.
Only when rollout evidence shows that candidates are semantically equivalent with the same preconditions and state effects may you prefer the lower-cost tool.
Never describe a cheaper, merely similar tool as a substitute.
Capture cost guidance only when evidence demonstrates an avoidable redundant call, an error/known-invalid retry, or an equivalent alternative.
Prefer the minimal correct workflow: authenticate once, obtain required IDs once, perform required writes, and verify or report only when the task requires it.

\medskip
\textbf{User Prompt:}

\textbf{Task Metadata:}

\{TASK\_METADATA\}

\medskip
\textbf{Reward Outcome:}

\{REWARD\}

\medskip
\textbf{Summarized Trajectory:}

\{TRAJECTORY\_SUMMARY\}

\medskip
\textbf{Planning Skills in Context:}

\{PLANNING\_SKILLS\}

\medskip
\textbf{Atomic Skills in Context:}

\{ATOMIC\_SKILLS\}

\medskip
\textbf{Tool Skills in Context:}

\{TOOL\_SKILLS\}

\end{tcolorbox}

\subsubsection{Retrieval-Policy Skill Distillation}
\label{app:retrieval_policy_prompt}

\begin{tcolorbox}[
    breakable,
    colback=black!5!white,
    colframe=black!60!black,
    left=1ex, right=1ex,
    top=0.5ex, bottom=0.5ex,
    fontupper=\small\ttfamily,
    fontlower=\small\ttfamily,
    title=\textbf{Retrieval-Policy Skill Distillation}
]
\textbf{System Prompt:}

You create concise retrieval-policy skills for an agent with a virtual \texttt{retrieve\_tools(query)} tool.

Use only the supplied evidence.
Write one conditional, reusable rule about when to retrieve, how to phrase a focused query, or how to assess retrieved schemas.
Do not name task-specific IDs, invent tools, or describe business-tool APIs.
Do not claim retrieval itself executes an action.

\medskip
\textbf{Output Format:}

Return exactly:
\begin{verbatim}
<policy>
{
  "name": "...",
  "document": "...",
  "content": "...",
  "policy_type":
    "query_reformulation|retrieval_control|result_assessment"
}
</policy>
\end{verbatim}

\textbf{Evidence:}

\medskip
\textbf{User Prompt:}

\{RETRIEVAL\_EVIDENCE\}

\end{tcolorbox}

\subsection{Initial Capability Provisioning}
\label{app:initial_provisioning_prompts}

\subsubsection{Task-Specific Pseudo Plan Construction}
\label{app:plan_construction_prompt}

\begin{tcolorbox}[
    breakable,
    colback=black!5!white,
    colframe=black!60!black,
    left=1ex, right=1ex,
    top=0.5ex, bottom=0.5ex,
    fontupper=\small\ttfamily,
    fontlower=\small\ttfamily,
    title=\textbf{Task-Specific Pseudo Plan Construction}
]
\textbf{System Prompt:}

You are rewriting retrieved planning skills into a task-specific pseudo-plan.
Use the current request and the retrieved plans.

\medskip
\textbf{Output Format:}

After thinking, return strict JSON only with the shape:
\begin{verbatim}
{"steps":["step 1","step 2"]}
\end{verbatim}

Each step must be a short actionable string.
Do not include commentary or any text outside the JSON.

\medskip
\textbf{User Prompt:}

\textbf{Current Request:}

\{QUERY\_TEXT\}

\medskip
\textbf{Retrieved Planning Skills:}

\{RETRIEVED\_PLANS\}

\end{tcolorbox}

\subsubsection{Skill Selection}
\label{app:skill_selection_prompt}

\begin{tcolorbox}[
    breakable,
    colback=black!5!white,
    colframe=black!60!black,
    left=1ex, right=1ex,
    top=0.5ex, bottom=0.5ex,
    fontupper=\small\ttfamily,
    fontlower=\small\ttfamily,
    title=\textbf{Skill Selection}
]
\textbf{System Prompt:}

Select the most applicable reusable skills for solving the current request.
Consider the request and pseudo-plan.

\medskip
\textbf{Output Format:}

Return strict JSON with the shape:
\begin{verbatim}
{"selected_indices":[1,2,...]}
\end{verbatim}

Select at most \{FINAL\_MAX\_SKILLS\} skills.

\medskip
\textbf{User Prompt:}

\textbf{Current Request:}

\{QUERY\_TEXT\}

\medskip
\textbf{Pseudo-plan:}

\{PSEUDO\_PLAN\}

\medskip
\textbf{Candidate Skills:}

\{CANDIDATE\_SKILLS\}

\end{tcolorbox}

\subsubsection{Skill Context Injection}
\label{app:skill_context_prompt}

\begin{tcolorbox}[
    breakable,
    colback=black!5!white,
    colframe=black!60!black,
    left=1ex, right=1ex,
    top=0.5ex, bottom=0.5ex,
    fontupper=\small\ttfamily,
    fontlower=\small\ttfamily,
    title=\textbf{Skill Context Injection}
]
\textbf{System Context Template:}

Reusable reference skills.
Use them only when applicable to the current request.
Any tools mentioned in these skills are only references and may not all be available in the current environment.
You must rely on the actual tool list provided in this session when deciding which tools can be called.

\medskip
\textbf{Selected Atomic Skills:}

\{ATOMIC\_SKILL\_BLOCKS\}

\medskip
\textbf{Tool Clarifications:}

\{TOOL\_CLARIFICATION\_BLOCKS\}

\end{tcolorbox}

\subsection{On-Demand Capability Provisioning}
\label{app:on_demand_provisioning_prompts}

\subsubsection{Capability Investigation Interface}
\label{app:capability_investigation_tool}

\begin{tcolorbox}[
    breakable,
    colback=black!5!white,
    colframe=black!60!black,
    left=1ex, right=1ex,
    top=0.5ex, bottom=0.5ex,
    fontupper=\small\ttfamily,
    fontlower=\small\ttfamily,
    title=\textbf{Capability Investigation Interface}
]
\textbf{Tool Name:}

\texttt{explore\_capabilities}

\medskip
\textbf{Tool Description:}

Delegate a focused investigation to the manager.
Use this when the currently available tools are clearly insufficient, when you cannot tell which tool expresses a specific constraint the user mentioned, or when a tool call failed because a required or unsupported parameter.
The manager searches the reusable skill library and the full cross-domain tool catalog, inspects tool schemas (up to \{MAX\_ACTIONS\} lookups), and returns a compact report plus a set of tools that become available to you.
When a stored procedure contains parameter-level or ordering detail, it can also add that procedure verbatim to your context under \texttt{\# Added skills}.
It never executes any tool, so it has no side effects and does not change any data.
Prefer \texttt{retrieve\_tools} for a cheap single keyword lookup; use this only for harder cases.

\medskip
\textbf{Parameters:}
\begin{itemize}
    \item \texttt{goal} (string, required):\\
    What you are trying to accomplish, in one or two sentences.
    Include the concrete user constraints you already know.

    \item \texttt{missing\_capability} (string, required):\\
    The specific capability or parameter you could not find with the tools you currently have.

    \item \texttt{known\_failures} (string, optional):\\
    Tools you already tried and why they did not work.
\end{itemize}

\end{tcolorbox}

\subsubsection{Tool Retrieval Interface}
\label{app:tool_retrieval_tool}

\begin{tcolorbox}[
    breakable,
    colback=black!5!white,
    colframe=black!60!black,
    left=1ex, right=1ex,
    top=0.5ex, bottom=0.5ex,
    fontupper=\small\ttfamily,
    fontlower=\small\ttfamily,
    title=\textbf{Tool Retrieval Interface}
]
\textbf{Tool Name:}

\texttt{retrieve\_tools}

\medskip
\textbf{Tool Description:}

Retrieve additional tools when the current tool list is insufficient.
Pass one short, specific natural-language query describing the tool capability you still need.
Use this only when the currently available tools do not cover the task.

\medskip
\textbf{Parameters:}
\begin{itemize}
    \item \texttt{query} (string, required):\\
    A short, specific description of the tool capability you want to retrieve.
\end{itemize}

\end{tcolorbox}

\subsubsection{On-Demand Capability Investigation}
\label{app:manager_investigation_prompt}

\begin{tcolorbox}[
    breakable,
    colback=black!5!white,
    colframe=black!60!black,
    left=1ex, right=1ex,
    top=0.5ex, bottom=0.5ex,
    fontupper=\small\ttfamily,
    fontlower=\small\ttfamily,
    title=\textbf{On-Demand Capability Investigation}
]
\textbf{System Prompt:}

You are the manager.
A main agent is solving a multi-turn task with tools, ran out of usable tools, and delegated a focused investigation to you.

You can search a reusable skill library, read skill bodies and walk the skill graph, search the full cross-domain tool catalog, inspect tool JSON schemas.
You CANNOT execute any real tool and you never talk to the user.
Your only deliverable is a short report plus the list of tools the main agent should be given.

\medskip
\textbf{Rules:}
\begin{enumerate}
    \item Be economical.
    You have at most \{MAX\_ACTIONS\} action calls in total.
    Start broad (\texttt{search\_tools} / \texttt{search\_skills}), then verify with \texttt{inspect\_tool}.
    You MAY request several independent actions in the same turn (e.g. inspect three candidate tools at once)---doing so is cheaper and strongly preferred over one action per turn.

    \item ALWAYS verify a tool with \texttt{inspect\_tool} before recommending it.
    Never invent a tool name or a parameter name: only names returned by the backend exist.

    \item Tools the main agent can already see are listed in the request.
    Do not recommend them again unless a parameter detail was the blocker.

    \item Tool names change across catalog versions.
    If a skill references a tool that no longer exists, search for a semantically equivalent replacement and say so explicitly.

    \item Prefer tools with lower cost / latency when several options are equivalent.

    \item Call \texttt{finish} as soon as you can answer the request---do not keep exploring just because you still have budget left.
\end{enumerate}

\medskip
\textbf{Addressing Skills (Important):}
\begin{enumerate}
    \setcounter{enumi}{6}
    \item Every skill card carries a \texttt{ref} such as
    \texttt{atomic:verify\_engineer} \texttt{\_credentials},
    \texttt{tool\_skill:record\_reading}, or
    \texttt{planning:1a2b3c4d}.
    \texttt{read\_skill} and \texttt{expand\_graph} accept ONLY that \texttt{ref}.
    Never pass a display name, a tool name, or a sentence copied from a card---it will not resolve.

    \item If a lookup fails you get \texttt{error} plus \texttt{did\_you\_mean}.
    Use one of those suggestions instead of retrying the same string.
    An empty \texttt{neighbors} list with \texttt{node\_found: true} means the node genuinely has no such edges---move on.

    \item When you need to know HOW to call a tool (required parameters, valid enum values, preconditions, ordering), use \texttt{search\_tool\_skills}.
    \texttt{inspect\_tool} gives you the schema; tool skills give you the learned usage rules.
\end{enumerate}

\medskip
\textbf{Reporting:}
\begin{enumerate}
    \setcounter{enumi}{9}
    \item Compress aggressively: the report goes into the main agent's context, so it must be short and actionable.
    Distill skills into imperative steps rather than pasting their bodies.

    \item If a skill body contains parameter-level or ordering detail that would be lost by paraphrasing, put its \texttt{ref} in \texttt{commit\_skills} (max \{MAX\_COMMIT\_SKILLS\})---the main agent will receive that procedure verbatim.
    Use it sparingly; it costs the main agent context.

    \item \texttt{commit\_skills} does NOT give the main agent the skill's tools.
    Any tool you want it to use must be inspected and listed in \texttt{commit\_tools}.
\end{enumerate}

\medskip
\textbf{Report Format:}

\texttt{finish(report=...)} must follow this exact layout (omit a section if empty):

\smallskip
\noindent
GOAL: <one sentence>\\
TOOLS:\\
\hspace*{1em}- <tool\_name>: <why it is needed> | required: <comma separated required params> | cost=<c> latency=<l>\\
WORKFLOW:\\
\hspace*{1em}1. <one line per step>\\
PITFALLS:\\
\hspace*{1em}- <one line per pitfall>\\
RULED\_OUT:\\
\hspace*{1em}- <tool\_name>: <why it does not work>

\medskip
\textbf{Hard Limits:}

At most \{MAX\_COMMIT\_TOOLS\} tools in TOOLS, at most 5 WORKFLOW steps, at most 3 PITFALLS, at most 3 RULED\_OUT entries, and at most \{MAX\_REPORT\_CHARS\} characters overall.

\medskip
\textbf{User Prompt:}

\textbf{GOAL:}

\{GOAL\}

\medskip
\textbf{MISSING CAPABILITY:}

\{MISSING\_CAPABILITY\}

\medskip
\textbf{KNOWN FAILURES:}

\{KNOWN\_FAILURES\}

\medskip
\textbf{RECENT USER MESSAGES:}

\{RECENT\_USER\_MESSAGES\}

\medskip
\textbf{RECENT TOOL ERRORS:}

\{RECENT\_TOOL\_ERRORS\}

\medskip
\textbf{PSEUDO-PLAN:}

\{PSEUDO\_PLAN\}

\medskip
\textbf{TOOLS ALREADY VISIBLE TO THE MAIN AGENT:}

\{VISIBLE\_TOOL\_NAMES\}

\medskip
\textbf{SKILLS ALREADY INJECTED:}

\{INJECTED\_SKILL\_NAMES\}

\end{tcolorbox}

\end{document}